%% file: latex/acl_latex.tex
\documentclass[11pt]{article}

\usepackage[preprint]{acl}

\usepackage{times}
\usepackage{latexsym}
\usepackage{booktabs}
\usepackage{multirow}
\usepackage{makecell}
\usepackage[table]{xcolor}
\usepackage{adjustbox}
\usepackage[T1]{fontenc}
\usepackage[utf8]{inputenc}

\usepackage{microtype}

\usepackage{inconsolata}

\usepackage{graphicx}
\usepackage{amsmath,amssymb,amsfonts}
\usepackage{algorithmic}
\usepackage{graphicx}
\usepackage{textcomp}
\usepackage{xcolor}
\usepackage[table]{xcolor}
\usepackage{booktabs}
\usepackage{multirow}
\usepackage{subcaption}
\title{SCOPE: Sequential Conformal Probing for Reliable OOD Rejection in LLM Services}

\author{
 \textbf{Zhuoyun Li},
 \textbf{Boxuan Wang},
 \textbf{Changshun Wu},
 \textbf{Xiaowei Huang},
 \textbf{Yi Dong\thanks{Corresponding Author: yi.dong@liverpool.ac.uk}}
 \\
 School of Computer Science and Informatics, University of Liverpool, United Kingdom
 \\
   \small{\texttt{\{zhuoyun.li, boxuan.wang, changshun.wu, xiaowei.huang, yi.dong\}@liverpool.ac.uk}}
}

\begin{document}
\maketitle
\begin{abstract}
Rejecting inputs outside the defined in-distribution (IND) service scope is critical for large language model (LLM) services, where unsupported requests should be filtered before full generation.
Existing out-of-distribution (OOD) detectors often rely on final outputs or final-layer representations, leaving unclear where service-boundary signals are most clearly encoded inside the model; they also lack a theoretical guarantee for held-out inputs.
In this paper, we introduce \textbf{SCOPE} (\textbf{S}equential \textbf{C}onformal \textbf{O}OD \textbf{P}robing and \textbf{E}valuation), a framework that selects a readable hidden layer, constructs a conformal gate with IND calibration, and uses a supermartingale e-process to certify persistent service-boundary evidence.
Experiments across multiple LLM backbones and six carefully designed boundary conditions show that SCOPE improves gate-level rejection over standard final-layer detectors, while revealing how different OOD boundaries take different geometric forms in hidden space.
\end{abstract}

\section{Introduction}
\label{sec:introduction}
Large language models (LLMs) are increasingly deployed as the front line of user-facing services, where deciding \emph{when not to answer} is as important as generating a fluent response. 
Consider an enterprise e-commerce assistant designed to handle order tracking and returns. 
If a user instead asks for medical advice, insurance-plan changes, or another request outside the supported service scope, the system should reject or reroute the input before full generation, rather than spending compute on unsupported traffic and risking an unreliable response~\cite{zhao2021detecting,dong2023reliability,jiang2025llmprism}. 
OOD rejection is therefore a service-level reliability problem: an effective gate should be lightweight, operate before generation when possible, and avoid unnecessarily dropping valid in-distribution (IND) requests. 
We study this problem for frozen LLM backbones, which are attractive in deployment because they offer cost-efficiency, controllability, and easy integration into existing service pipelines~\cite{touvron2023llama2openfoundation,qwen2025qwen25technicalreport,li2025sr}.

\begin{figure}
    \centering
    \includegraphics[width=1\columnwidth]{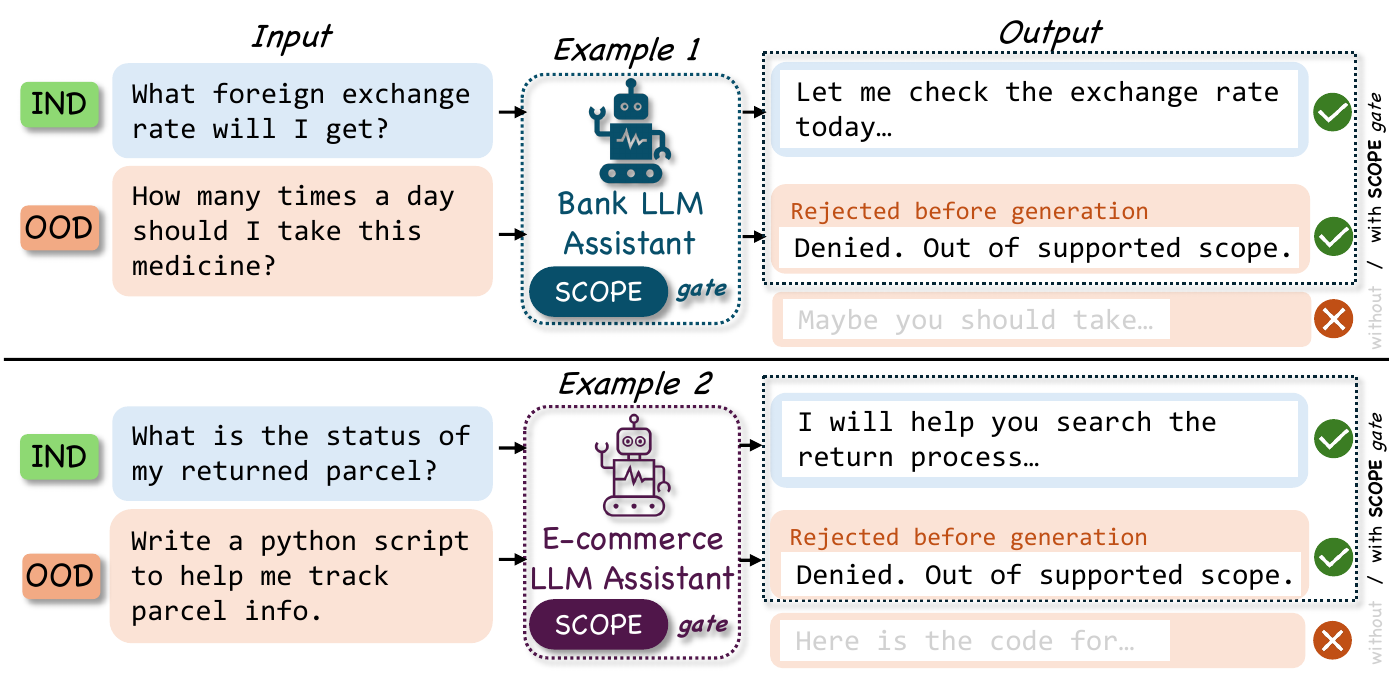}
    \caption{\textbf{Examples for Service-scope OOD rejection before generation.}
A task-specific LLM service should answer inputs within its IND scope and reject or route unsupported inputs before full generation. SCOPE provides a reliable gate for pre-generation rejection.}
    \label{fig:SCOPEexample}
\end{figure}
Existing OOD detectors provide useful building blocks, but a deployment-time LLM rejection gate requires more than an offline score. 
Current approaches generally follow three paradigms. 
Output-based methods score the final probability distribution or use generation consistency to estimate uncertainty~\cite{hendrycks2017a,NEURIPS2020_f5496252,kuhn2023semantic,farquhar2024semantic}. 
Representation-based methods use distance or density estimates in hidden states, often at the final layer~\cite{NEURIPS2018_abdeb6f5}. 
Calibration-oriented methods provide post-hoc thresholds or conformal guarantees for a chosen score~\cite{Lei03072018,novello2024outofdistributiondetectionuseconformal,gupta-etal-2025-polysemantic}. 
These components leave three gaps for LLM service gates. 
First, final-output and final-layer methods assume that the relevant OOD signal is best exposed at the end of the model, although domain, intent, and rewrite-induced shifts may appear at different depths and become entangled near the output layer~\cite{alain2017linearprobes,gurnee2024spacetime}. 
Second, consistency-based uncertainty often requires multiple generations, which is poorly matched to a lightweight pre-generation gate. 
Third, calibration fixes an operating threshold, but does not by itself test whether calibrated rejections persist as valid evidence on a held-out stream.

To address these gaps, we introduce \textbf{SCOPE} (\textbf{S}equential \textbf{C}onformal \textbf{O}OD \textbf{P}robing and \textbf{E}valuation), a representation-relative framework for service-boundary certification. 
It searches across transformer layers on disjoint development data and selects the layer where the service-boundary signal is most readable. 
It then fits a lightweight linear readout at the selected layer and calibrates its rejection threshold using only IND calibration data. 
We call this main instantiation the \emph{Conformal Linear Gate} (CLG), which provides an explicit finite-sample budget on IND false rejections while leaving the LLM frozen. 
Finally, SCOPE constructs a nonnegative supermartingale e-process on top of the conformalized rejection decisions~\cite{shafer2011test,howard2020time,grunwald2024beyond}. 
This turns repeated calibrated rejections into anytime-valid evidence that the fixed gate exposes a persistent service-boundary signal on a held-out stream.

We evaluate SCOPE across multiple LLM backbones and six IND/OOD service-boundary conditions, ranging from far-OOD transfer to near-domain shifts, fine-grained intent boundaries, paraphrase stress, and same-distribution null streams. In summary, our main contributions are threefold: \textbf{First}, we formulate OOD rejection for LLM services as a representation-relative service-boundary certification problem, testing whether a frozen model exposes a stable and readable boundary signal at a selected layer.
\textbf{Second}, we introduce a framework \textbf{SCOPE} to build calibrated and anytime-valid OOD rejection gates. Its main instantiation, the Conformal Linear Gate (CLG), provides a lightweight front-end without modifying the LLM.
\textbf{Third}, we conduct a multi-model, multi-boundary evaluation showing that SCOPE gives better behavior than standard detectors, while revealing how different OOD boundaries correspond to different selected-layer signal geometries.

\begin{figure*}[t]
    \centering
    \includegraphics[width=1\linewidth]{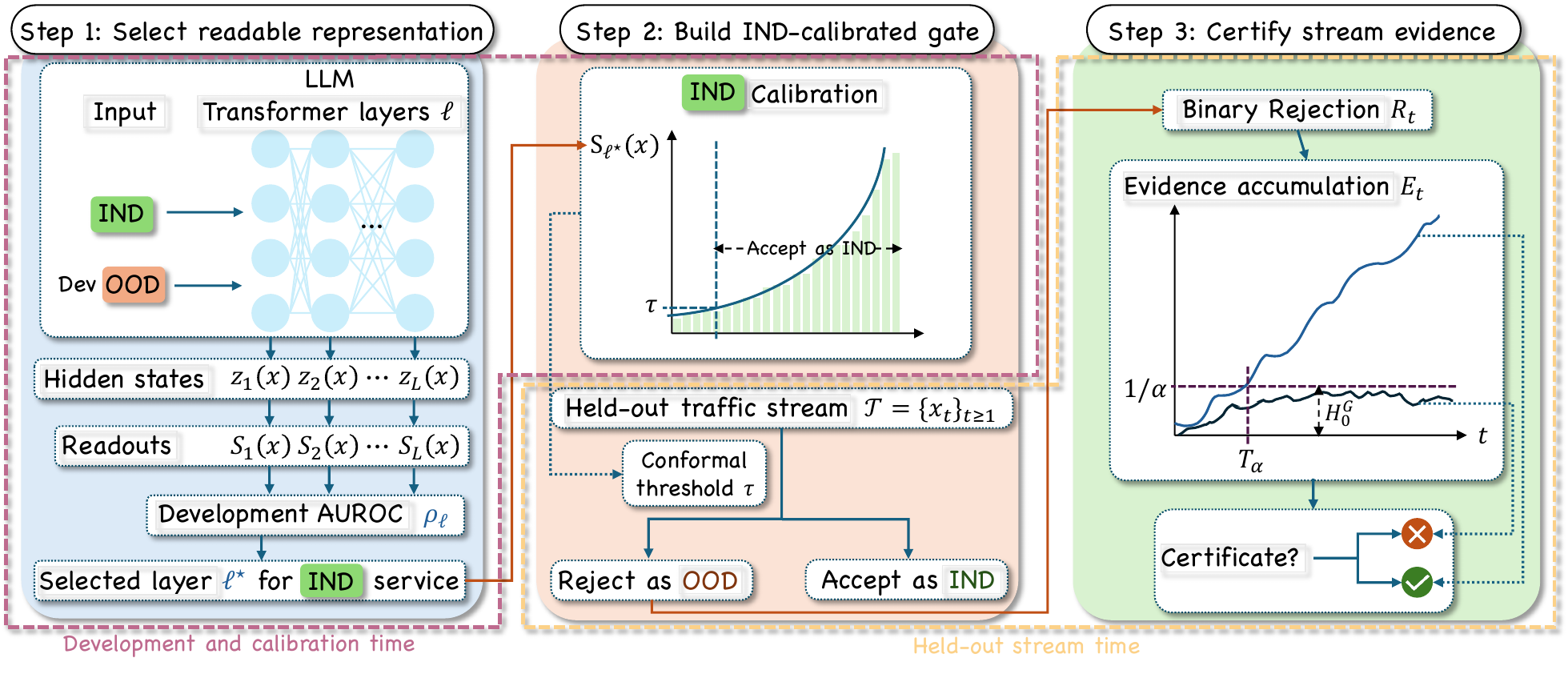}
    \caption{
\textbf{Overview of SCOPE. }
The framework first selects a readable hidden layer from frozen LLM representations, then calibrates the selected score into an IND-controlled rejection gate, and finally accumulates the held-out rejection stream with an e-process to obtain an anytime-valid service-boundary certificate.
}
    \label{fig:method_overview}
\end{figure*}

% \subsection{Calibration and Sequential Testing}

% Conformal prediction provides distribution-free calibration with finite-sample validity under exchangeability. Split conformal methods are especially useful in our setting because they turn arbitrary real-valued scores into calibrated acceptance or rejection rules using a held-out calibration set, avoiding heuristic threshold tuning while controlling false rejections on future IND inputs~\cite{Lei03072018}. Recent work further studies conditional or subgroup validity and the design of practical nonconformity scores within conformal pipelines~\cite{gibbs2025conformal,pmlr-v204-kato23a}. Our sequential component is related to test martingales, e-values, e-processes, and nonnegative supermartingales for anytime-valid inference. These tools provide evidence measures that remain valid under optional stopping and continuous monitoring~\cite{shafer2011test,howard2020time,grunwald2024beyond}. While conformal calibration and sequential testing are usually developed separately from representation-level probing in LLMs, our framework connects them in a single OOD rejection pipeline: a layerwise probe provides the score, split conformal calibration controls IND false rejection, and a supermartingale e-process tests whether the selected representation separates IND from OOD inputs over time.

\section{Related Work}
\label{sec:relatedwork}

\subsection{OOD Detection and Probing for LLMs}

Reliable rejection for LLM services draws on work on uncertainty, abstention, OOD detection, and representation probing. Language models can often report whether they know an answer, motivating abstention-style handling of unknown or unsupported queries~\cite{kadavath2022know}. Related studies examine semantic uncertainty, calibrated ``I don't know'' behavior, truthfulness, and hallucination detection in generation~\cite{kuhn2023semantic,farquhar2024semantic,deng2024dontknow,bayat2024lito,jiang-etal-2024-large,ahdritz2024knowable,wang2023hallucination}. More direct OOD work evaluates LLM-based representations and lightweight detectors for near-OOD intent recognition and broader LLM OOD detection settings~\cite{sali2025navigating,liu-etal-2024-good}, while multimodal work explores OOD detection through LLM-generated class descriptions or uncertainty over vision-language representations~\cite{dai2023exploring,li2024referencefree}. These studies provide important reliability signals, but LLM service deployment also requires a lightweight pre-generation gate that decides whether an input is supported by the intended IND service scope.
A parallel line of work uses probing to inspect what information is available in neural representations. Early work measures layerwise linear separability in neural networks~\cite{alain2017linearprobes}. Recent LLM studies show that truthfulness, correctness, uncertainty, spatial-temporal structure, and other semantic attributes can often be recovered with simple linear readouts~\cite{azaria2023internal,marks2023geometry,pmlr-v235-park24c,gurnee2024spacetime,bouchaud2025linear,bao-etal-2025-probing,cencerrado2025noanswer,deng2024dontknow,ahdritz2024knowable}. SCOPE uses this probing perspective for \emph{distributional support}: it searches across transformer layers of a frozen LLM for the representation where an IND/OOD service boundary is most readable, then turns the selected readout into a calibrated rejection gate with sequential evidence accumulation.

\subsection{Calibration and Sequential Testing}

Conformal prediction provides distribution-free calibration with finite-sample validity under exchangeability~\cite{Lei03072018}. Split conformal methods are well suited to service rejection because they convert any fixed real-valued score into an acceptance or rejection rule using held-out IND calibration data, giving an explicit budget on IND false rejections without heuristic threshold tuning. Recent work further studies conditional or subgroup validity and practical nonconformity scores within conformal pipelines~\cite{gibbs2025conformal,pmlr-v204-kato23a}. In SCOPE, conformal calibration serves as the operating interface for a frozen selected-layer OOD score.
Our stream-level certificate is related to test martingales, e-values, e-processes, and nonnegative supermartingales for anytime-valid inference~\cite{shafer2011test,howard2020time,grunwald2024beyond}. These tools provide evidence measures that remain valid under optional stopping and continuous monitoring, which matches deployed services where traffic arrives sequentially. SCOPE connects this sequential testing view to representation-level LLM probing: a layerwise readout provides the score, IND-only conformal calibration fixes the rejection threshold, and a supermartingale e-process tests whether the fixed gate rejects a held-out stream persistently above the IND-calibrated baseline.

\section{Method}
\label{sec:method}

SCOPE turns LLM representations into a service-level OOD gate in three steps as Figure~\ref{fig:method_overview}.
It first selects a hidden layer where the service boundary is readable, then calibrates the selected score into an IND-controlled rule, and finally uses an e-process to certify whether the resulting rejection stream carries persistent service-boundary evidence.

\subsection{Selecting a Readable Representation}
\label{sec:select_layer}

We consider a frozen LLM \(f_\theta\) used as the backbone of a deployed service. For an input \(x\), let

{\small \[
z_\ell(x) := Z^{f_\theta}_\ell(x) \in \mathbb{R}^d
\]}
denote the last-token hidden representation at transformer layer \(\ell\). Each example has a service-side domain label \(D\in\{0,1\}\), where \(D=0\) denotes in-distribution (IND) traffic and \(D=1\) denotes out-of-distribution (OOD) traffic.

The certificate is defined for a fixed gate, specified by the frozen model, the selected layer, the scalar score rule, the conformal threshold, and the e-process parameter. In this view, boundary separability is model-induced: the same IND/OOD split may be clearly exposed by one backbone and only weakly exposed by another.

We keep the data roles separated throughout. The IND data are split into \(\mathcal I_d\), \(\mathcal I_c\), and \(\mathcal I_e\), used for development, conformal calibration, and held-out IND evaluation, respectively. The development IND split \(\mathcal I_d\) supplies the IND side for both readout fitting and layer selection. A representative development OOD source is split into \(\mathcal O_r\) and \(\mathcal O_s\), used for readout fitting and layer selection. The final held-out OOD or null stream is denoted by \(\mathcal T\). The layer, score rule, conformal threshold, and e-process parameter are fixed before \(\mathcal T\) is read.

The method only requires a scalar OOD score \(S(x)\) fixed before calibration and stream evaluation. We use a linear hidden-state readout as the main score module. This gives a direct certification interface: when a low-capacity readout supports a certificate, the service-boundary signal is accessible from the frozen representation. The readout is also inexpensive, works on cached hidden states, and leaves the LLM unchanged.

Let \(m\) index the score family. In our main construction, \(m\in\{\mathrm{lin},\mathrm{dir}\}\), corresponding to a full linear readout and a one-dimensional directional readout. For each layer \(\ell\), we fit a layerwise score \(S_{\ell,m}\) using \(\mathcal I_d\) and \(\mathcal O_r\).

\paragraph{Linear readout.}
For layer \(\ell\), let \(w_\ell\in\mathbb R^d\) and \(b_\ell\in\mathbb R\) be the weight vector and bias of a logistic readout fitted on \(\mathcal I_d\) and \(\mathcal O_r\). With \(\sigma(\cdot)\) denoting the logistic sigmoid, the layerwise linear OOD score is

{\small \[
S_{\ell,\mathrm{lin}}(x)
=
\sigma\!\left(w_\ell^\top z_\ell(x)+b_\ell\right).
\]}

Larger values indicate that the layer-\(\ell\) representation of \(x\) is more OOD-like under the fitted readout.

\paragraph{Directional readout.}
As a lower-capacity companion, we also define a rank-one directional score. Let \(\mu_\ell^{\rm IND}\) and \(\mu_\ell^{\rm OOD}\) be the mean representations of \(\mathcal I_d\) and \(\mathcal O_r\) at layer \(\ell\).
The directional score is

{\small \[
S_{\ell,\mathrm{dir}}(x)
=
g_\ell\!\left(d_\ell^\top
(z_\ell(x)-\mu_\ell^{\rm IND})\right),
\]}

where 
{\small 
$d_\ell
=
(\mu_\ell^{\rm OOD}-\mu_\ell^{\rm IND})/
(\|\mu_\ell^{\rm OOD}-\mu_\ell^{\rm IND}\|_2)$}, and \(g_\ell\) is a monotone map fitted on development data so that larger values indicate stronger OOD evidence.

\paragraph{Layer selection.}
For each score family \(m\), we choose the layer by development AUROC:

{\small \[
\begin{aligned}
\rho_{\ell,m}
&=
\operatorname{AUROC}(S_{\ell,m};\mathcal I_d,\mathcal O_s),\\
\ell_m^\star
&=
\arg\max_{\ell\in\{1,\ldots,L\}}\rho_{\ell,m},
\end{aligned}
\]}

where \(L\) is the number of transformer layers and \(\mathcal O_s\) is the held-out development OOD split. After this step, the selected pair \((\ell_m^\star,S_{\ell_m^\star,m})\) is frozen. When the score family is clear from context, we write \(\ell^\star\) and \(S\).

\subsection{Building an IND-Calibrated Rejection Gate}
\label{sec:conformal_gate}

Given the fixed score \(S\), we calibrate the rejection threshold using an IND-only calibration set
\(\mathcal C_{\rm IND}=\{x_i^{\rm cal}\}_{i=1}^{M}\subset\mathcal I_c\).
For a target IND false rejection budget \(\varepsilon\in(0,1)\), we set \(\tau\) to the standard split-conformal \((1-\varepsilon)\)-quantile of the calibration scores \(\{S(x_i^{\rm cal})\}_{i=1}^{M}\). The gate is

{\small \[
\phi(x)=\mathbf 1\{S(x)>\tau\},
\]}
where \(\phi(x)=1\) means that the input is rejected.

Under exchangeability between \(\mathcal C_{\rm IND}\) and future IND traffic, the IND false rejection probability

{\small \[
\mathbb P_{x\sim{\rm IND}}\bigl(S(x)>\tau\bigr)
\leq
\varepsilon+\frac{1}{M+1}.
\]}
We write
{\small \[
\bar\varepsilon
:=
\varepsilon+\frac{1}{M+1}.
\]}
This quantity is the finite-sample IND rejection baseline used by the sequential test. OOD rejection is evaluated as power on held-out streams.

We call the selected-layer linear version of this calibrated gate the \emph{Conformal Linear Gate} (CLG). The directional version is the \emph{Conformal Directional Gate} (CDG).

\subsection{Certifying Service-Boundary Evidence}
\label{sec:eprocess}

The calibrated gate maps a held-out stream \(\mathcal T=\{x_t\}_{t\geq 1}\) into binary rejections
\(R_t=\mathbf 1\{S(x_t)>\tau\}\).
We test if these rejections occur persistently above the IND baseline. 
The operational null is

{\small \[
% \begin{aligned}
H_0^{G}:\quad
q_t
:=
\mathbb P_{H_0}(R_t=1\mid\mathcal F_{t-1})\\
\leq
\bar\varepsilon,
\quad \text{for all }t ,
% \end{aligned}
\]}
where \(\mathcal F_{t-1}\) is the history before observing \(x_t\). This null states that the fixed gate rejects the stream no more often than the IND-calibrated baseline.

Before reading \(\mathcal T\), we fix an alternative rejection rate \(p_1\in(\bar\varepsilon,1)\), which may be chosen as a design constant or from development data, but is not updated on the held-out stream. 
Given \(p_1\), define

{\small \[
% \begin{aligned}
e_t
=
\left(\frac{p_1}{\bar\varepsilon}\right)^{R_t}
\left(\frac{1-p_1}{1-\bar\varepsilon}\right)^{1-R_t},
E_t
=
\prod_{i=1}^{t}e_i,
\quad E_0=1.
% \end{aligned}
\]}
Under \(H_0^{G}\), because \(q_t\leq\bar\varepsilon\) and \(p_1>\bar\varepsilon\),

{\small \[
\mathbb E_{H_0}[e_t\mid\mathcal F_{t-1}]
=
q_t\frac{p_1}{\bar\varepsilon}
+
(1-q_t)\frac{1-p_1}{1-\bar\varepsilon}
\leq 1.
\]}
Therefore, \((E_t)_{t\geq0}\) is a nonnegative supermartingale under \(H_0^{G}\).
For significance level \(\alpha\), define
{\small \[
% \begin{aligned}
T_\alpha
=
\inf\{t\geq1:E_t\geq1/\alpha\},\
\mathbb P_{H_0}
\left(
\sup_{t\geq1}E_t\geq\frac1\alpha
\right)
\leq
\alpha ,
% \end{aligned}
\]}
where the probability bound follows from Ville's inequality. Thus, after the gate and \(p_1\) are fixed, the probability of falsely certifying service-boundary signal is at most \(\alpha\), even under sequential inspection.
Crossing \(1/\alpha\) certifies that the fixed selected-layer gate rejects above the IND-calibrated baseline, and \(T_\alpha\) records the number of stream samples needed to reach this evidence level. If the threshold is not crossed within the evaluation budget, the stream provides insufficient evidence for this fixed model, layer, and score rule.

\section{Experiment}
\label{sec:experiment}

We evaluate whether the proposed conformalized selected-layer gate can serve as a reliable OOD rejection front-end for frozen LLM services. 
The experiments are organized around three questions. 
\textbf{RQ1}: Does the proposed gate outperform standard final-output or final-layer OOD detectors under a controlled protocol? 
\textbf{RQ2}: Under which IND/OOD service boundaries does the calibrated gate produce stable sequential evidence? 
\textbf{RQ3}: Why do different boundary definitions lead to different rejection behavior?

% These questions map directly to the three parts of our evaluation. 
% Section~\ref{sec:controlled_baseline} compares the proposed gate with conventional OOD detectors on a fixed backbone. 
% Section~\ref{sec:cross_model_rejection} applies the same gate across model families and six service-boundary conditions. 
% Section~\ref{sec:ood_signal_geometry} analyzes the geometry of selected-layer OOD signals, explaining why some boundaries are naturally separable while others depend on the type of shift being read from the representation.

\subsection{Experimental Setup}
\label{sec:exp_setup}

\paragraph{Models.}
We use \textsc{LLaMA2}-7B~\citep{touvron2023llama2openfoundation} as the controlled backbone for the main detector comparison, so that the model, data split, and evaluation protocol are fixed while only the OOD detector varies. 
For the cross-model study, we further evaluate \textsc{Qwen2.5}-1.5B, \textsc{Qwen2.5}-7B, \textsc{Qwen2.5}-14B~\citep{qwen2025qwen25technicalreport}, \textsc{Mistral}-7B~\citep{jiang2023mistral7b}, \textsc{OLMo}-2-7B~\citep{olmo2025olmo2}, and \textsc{Falcon}-7B~\citep{almazrouei2023falcon}. 
All LLMs are frozen and used in inference-only mode; only lightweight detectors on cached hidden representations are trained.

\paragraph{IND domains and service boundaries.}
We use four IND domains: SST-2~\citep{socher2013recursive,wang2018glue} for sentiment, 20 Newsgroups~\citep{lang1995newsweeder} for topic classification, and CLINC150~\citep{larson2019evaluation} and Banking77~\citep{casanueva2020efficient} for intent classification. 
The evaluation covers six service-boundary conditions: P1 uses SST-2 as IND, RTE~\citep{dagan2006pascal,wang2018glue} as development OOD, and WMT14 De-En~\citep{bojar2014findings} as held-out far-OOD; P2 uses 20 Newsgroups as IND, TREC~\citep{li2002learning} as development OOD, and MNLI~\citep{williams2018broad} as held-out cross-task OOD; P3 uses CLINC150 Travel as IND and CLINC150 Banking as near-domain OOD; P4 uses a fixed Banking77 intent split with 38 IND intents and 39 OOD intents; P5 uses the same Banking77 IND side as P4 but evaluates 500 intent-preserving paraphrases generated by \textsc{Qwen2.5}-14B; and P6 is a same-distribution null stream, using disjoint held-out SST-2 examples. 
Thus, P1--P3 cover broad or moderate OOD transfer, P4 tests a fine-grained intent boundary, P5 tests rewrite-induced shift under preserved intent, and P6 checks false certification on IND traffic.
For ease of reference, Appendix~\ref{app:boundary_construction} summarizes P1--P6 in a compact table.

\paragraph{Splits and layer selection.}
For each model--boundary configuration, we run five random seeds. 
The semantic boundary definition is fixed across seeds; the seed only affects the internal split and stream order. 
For each seed, the IND pool is split into 70\% probe-training data, 10\% conformal-calibration data, and 20\% held-out IND test data. 
The representative OOD source is split into 70\% probe-training data and 30\% held-out development data. 
Layerwise probes are trained on the probe-training splits, the best layer $\ell^\star$ is selected by development AUROC, and this layer is frozen before held-out stream evaluation. 
The test OOD or null stream is never used for probe fitting, conformal calibration, layer selection, threshold tuning, or e-process parameter selection.

\paragraph{Metrics.}
We report AUROC and FPR@95 for offline OOD ranking. 
After conformal calibration, we report empirical IND false rejection and OOD-TPR@$\tau$, which measure the calibrated operating point and held-out OOD rejection rate. 
For the e-process, certificate rate records the fraction of shuffled held-out streams crossing $1/\alpha$, and $T_\alpha$ records the stopping time when certification succeeds. 
For each seed, we run 20 shuffled streams using the fixed held-out decisions; shuffling changes only arrival order and does not alter training, calibration, layer selection, or test data.

\subsection{Controlled Detector Comparison and Gate Validation}
\label{sec:controlled_baseline}

This section answers \textbf{RQ1}: whether the proposed selected-layer conformal gate gives a stronger service-level operating point than standard OOD detectors. 

We use \textsc{LLaMA2}-7B as a controlled backbone and keep the model, split, and evaluation protocol fixed across all methods. 
This isolates the detector design: a useful gate should rank OOD examples highly while maintaining low IND false rejection at high OOD recall.

Table~\ref{tab:controlled_detector} shows that selected-layer CLG is the strongest detector in this controlled comparison. 
It achieves near-perfect AUROC on SST2, CLINC-Banking, and CLINC-Travel, remains substantially stronger on the harder Banking77 split, and gives the lowest FPR@95 across all five settings. 
This matters for deployment because a score with good average ranking is not sufficient if high OOD recall requires rejecting many valid IND examples. 
Mahalanobis performs well on far-OOD pairs but becomes unstable on intent-level settings, while MSP and energy vary sharply across domains. 
CLG therefore provides the most stable default operating point.
The comparison between CDG and CLG further clarifies the role of readout capacity. 
CDG performs well in easier settings, indicating that the selected layer often contains a simple OOD-sensitive direction. 
However, it drops on tighter intent boundaries, where a one-dimensional direction is not expressive enough. 
CLG keeps the same lightweight selected-layer design but learns a supervised linear boundary, giving a stronger readout without modifying the frozen LLM.

\begin{table}[t]

\centering
\small
\setlength{\tabcolsep}{4.0pt}
\renewcommand{\arraystretch}{1.12}

\definecolor{propbg}{RGB}{232,245,233}

\begin{adjustbox}{max width=\columnwidth}
\begin{tabular}{llccccc}
\toprule
\textbf{Method} & \textbf{Metric}
& \textbf{SST2}
& \textbf{20NG}
& \makecell{\textbf{Clinc}\\\textbf{Bank}}
& \makecell{\textbf{Clinc}\\\textbf{Travel}}
& \textbf{Banking77} \\
\midrule

\multirow{2}{*}{Maha}
& AUROC $\uparrow$
& 0.9991 & \textbf{0.9997} & 0.7985 & 0.8609 & 0.6713 \\
& FPR@95 $\downarrow$
& 0.0048 & 0.0018 & 0.9633 & 0.7450 & 0.8263 \\
\addlinespace[0.25em]

\multirow{2}{*}{Cosine}
& AUROC $\uparrow$
& 0.9957 & 0.8657 & 0.5335 & 0.6358 & 0.4180 \\
& FPR@95 $\downarrow$
& 0.0277 & 0.7031 & 0.9567 & 0.9650 & 0.9974 \\
\addlinespace[0.25em]

\multirow{2}{*}{MSP}
& AUROC $\uparrow$
& 0.6355 & 0.7993 & 0.8895 & 0.9115 & 0.4684 \\
& FPR@95 $\downarrow$
& 0.9373 & 0.6684 & 0.5000 & 0.4100 & 0.9699 \\
\addlinespace[0.25em]

\multirow{2}{*}{Energy}
& AUROC $\uparrow$
& 0.5603 & 0.7073 & 0.8874 & 0.9200 & 0.4180 \\
& FPR@95 $\downarrow$
& 0.9812 & 0.6983 & 0.5600 & 0.4850 & 0.9974 \\
\addlinespace[0.25em]

\multirow{2}{*}{CDG}
& AUROC $\uparrow$
& 0.9973 & 0.9767 & 0.8039 & 0.8092 & 0.7004 \\
& FPR@95 $\downarrow$
& 0.0283 & 0.0260 & 0.5796 & 0.6593 & 0.8055 \\
\midrule

\rowcolor{propbg}
\textbf{CLG (ours)}
& AUROC $\uparrow$
& \textbf{1.0000} & 0.9858 & \textbf{0.9999} & \textbf{1.0000} & \textbf{0.9582} \\

\rowcolor{propbg}
% \textbf{(ours)}
& FPR@95 $\downarrow$
& \textbf{0.0000} & \textbf{0.0000} & \textbf{0.0000} & \textbf{0.0000} & \textbf{0.1816} \\

\bottomrule
\end{tabular}
\end{adjustbox}
\caption{\textbf{Controlled detector comparison on main backbone.}
All methods use the same data splits and evaluation protocol. 
CLG gives the strongest overall gate-level operating point.}
\label{tab:controlled_detector}
\end{table}

\begin{figure}[t]
    \centering
    \includegraphics[width=0.85\columnwidth]{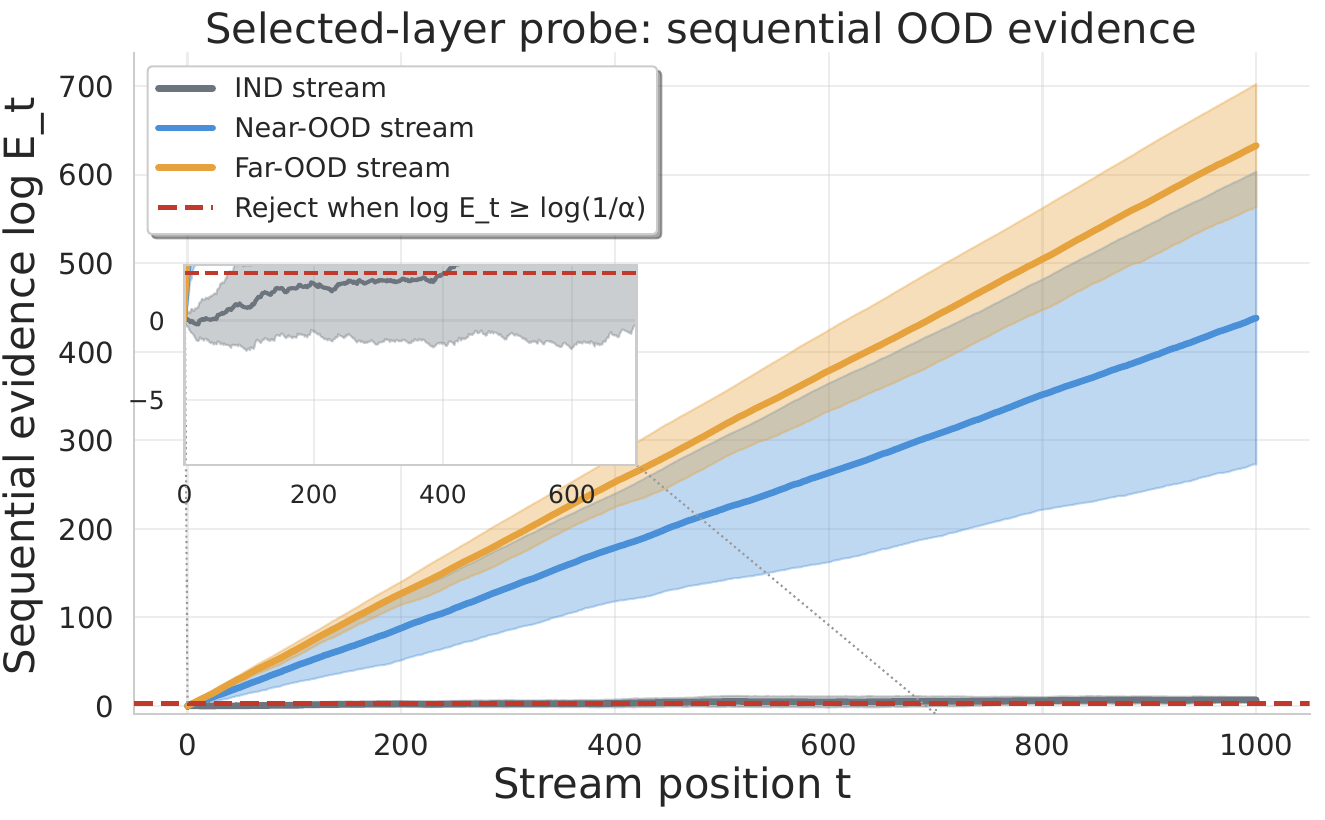}
    \caption{\textbf{Sequential evidence from the selected-layer gate.}
    After conformal calibration, CLG produces binary rejection decisions that are accumulated by the e-process. Far-OOD streams cross the anytime-valid threshold quickly, near-OOD streams accumulate evidence more gradually, and the IND-only stream stays below the threshold.}
    \label{fig:eprocess_logreg}
\end{figure}

Figure~\ref{fig:eprocess_logreg} shows the same gate after conformal calibration. 
The score is converted into binary rejection decisions and accumulated by the e-process. 
Far-OOD traffic accumulates evidence rapidly, near-OOD traffic grows more gradually, and the IND-only stream stays below the certification threshold. 
Thus, the certificate is driven by repeated IND-calibrated rejections rather than by an uncalibrated confidence score.

Together, Table~\ref{tab:controlled_detector} and Figure~\ref{fig:eprocess_logreg} answer RQ1. 
CLG improves the detector operating point over standard baselines, fixes the rejection threshold through IND-side conformal calibration, and converts persistent OOD rejections into sequential evidence. 
We therefore use CLG as the default gate in the cross-model boundary audit.

\subsection{Cross-Model Service-Boundary Certification}
\label{sec:cross_model_rejection}

This section answers \textbf{RQ2}: under which IND/OOD service boundaries the calibrated gate produces stable sequential evidence. 
We apply the same selected-layer CLG across seven frozen LLM backbones and six boundary conditions. 
For each model--boundary pair, the layer, score, conformal threshold, and e-process parameters are fixed before the held-out stream is evaluated. 
Thus, the heatmap does not merely report offline OOD ranking; it tests whether a development-selected gate continues to produce calibrated rejection evidence on unseen traffic.

\begin{figure}[t]
\centering
\includegraphics[width=1\columnwidth]{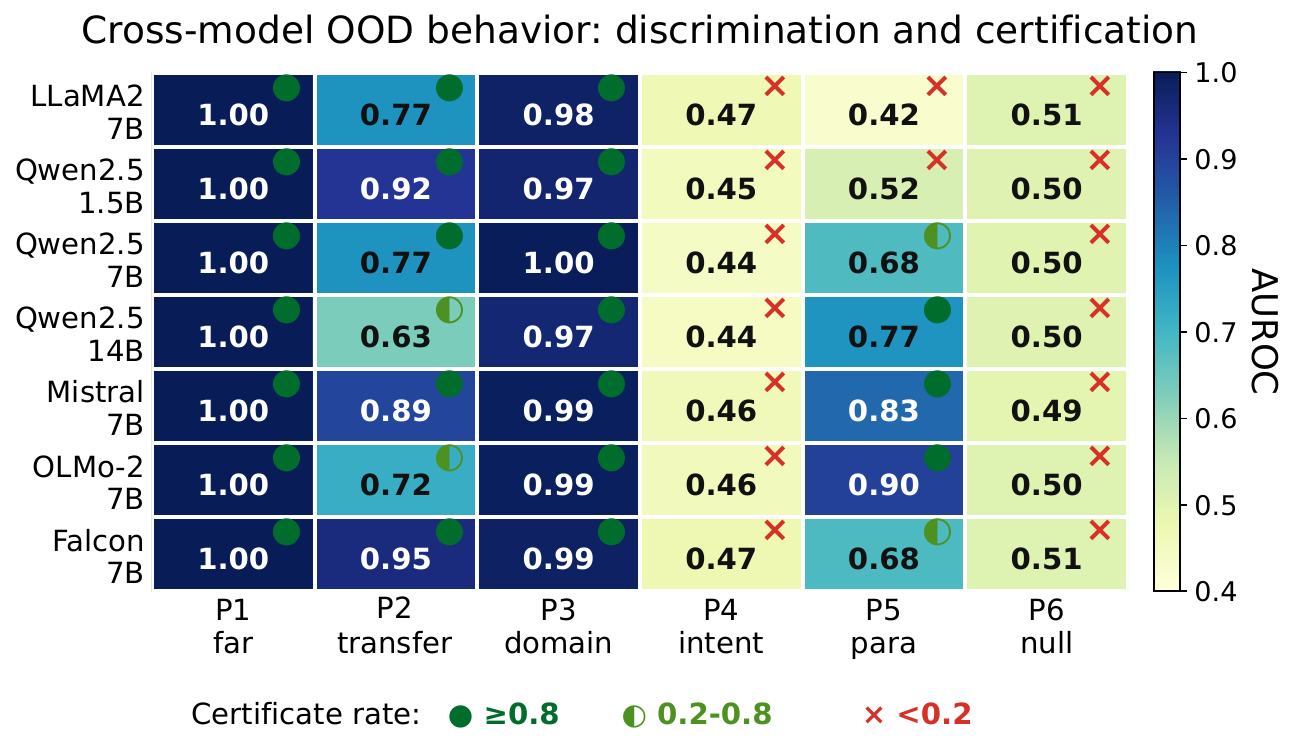}
\caption{\textbf{Cross-model service-boundary certification.}
Each cell reports CLG on one backbone and one boundary. 
Color shows AUROC; markers show e-process certificate rate over shuffled held-out streams.}
\label{fig:crossmodel_overview}
\end{figure}

Figure~\ref{fig:crossmodel_overview} shows a clear hierarchy of service boundaries. 
We mark certificate rate $\geq 0.8$ as reliable, $0.2$--$0.8$ as partial, and $<0.2$ as weak certification. 
These markers summarize stability across random splits and stream orders; the anytime-valid decision for each individual stream is still determined by whether the e-process crosses $1/\alpha$.

The strongest certification appears on broad and moderate OOD shifts. 
P1 is the clearest far-OOD transfer case: all seven backbones obtain near-perfect AUROC and reliable certificate rates. 
This shows that when the service boundary is broad, the selected intermediate layer exposes a strong OOD signal that survives both IND-only conformal calibration and sequential testing. 
P3 is more informative because CLINC150 Banking is closer to CLINC150 Travel than the far-OOD tasks in P1. 
Most backbones still achieve high AUROC and reliable certification, showing that the gate can certify domain-level shifts within a related intent benchmark rather than only trivial dataset mismatch.

P2 tests whether the selected layer transfers beyond the representative development OOD source. 
Here the gate is selected using TREC and evaluated on MNLI. 
The generally high AUROC indicates that the selected-layer score transfers to the held-out OOD task, while the variation in certificate rate shows why AUROC alone is not enough. 
A score may rank OOD examples above IND examples on average, but certification requires the held-out stream to cross the IND-calibrated threshold repeatedly enough to accumulate e-process evidence. 
Thus, P2 highlights the added value of evaluating calibrated rejection and sequential evidence in addition to offline ranking.

P4 and P5 define tighter service boundaries. 
P4 is a fixed Banking77 intent-level split, where IND and OOD examples are semantically close customer intents from the same benchmark. 
Across models, both AUROC and certificate rate are much lower than in P1--P3, indicating that this boundary is substantially finer than far-OOD or domain-level transfer. 
P5 is different: the inputs preserve the original Banking77 intent but are rewritten as paraphrases. 
Several backbones show moderate AUROC or partial certification, suggesting that rewrite-induced distribution shift is visible in some selected-layer representations. 
Because P5 preserves intent, its interpretation depends on the service definition; we analyze this boundary effect more directly in Sec.~\ref{sec:ood_signal_geometry}.

P6 is the same-distribution null stream. 
It uses held-out SST-2 examples that are disjoint from probe training, layer selection, and conformal calibration. 
AUROC stays near chance and certificate rates remain low across backbones. 
This check supports the validity of the certificates observed in P1--P3: the e-process is not triggered by repeated thresholding itself, but by persistent departures from the IND-calibrated rejection baseline.

Overall, Figure~\ref{fig:crossmodel_overview} answers RQ2 by showing that CLG produces stable sequential evidence on broad far-OOD and domain-level shifts, transfers nontrivially from development OOD to held-out OOD tasks, and remains controlled on same-distribution traffic. 
The contrast between P4 and P5 further shows that service-boundary definition is not a secondary detail: fine-grained intent splits and intent-preserving rewrites induce different hidden-state signals. 
This motivates the readout-geometry analysis in Sec.~\ref{sec:ood_signal_geometry}.

\begin{table*}[t]
\centering
\scriptsize
\setlength{\tabcolsep}{1.2pt}
\renewcommand{\arraystretch}{1.15}

\newcommand{\metric}[2]{$#1\;\text{\scriptsize($\pm$#2)}$}
\newcommand{\bmetric}[2]{$\mathbf{#1}\;\text{\scriptsize($\pm$#2)}$}

\begin{tabular}{@{}lccccccccc@{}}
\toprule
\multirow{2}{*}{\textbf{Readout}}
& \multicolumn{3}{c}{\textbf{P1 far}}
& \multicolumn{3}{c}{\textbf{P2 transfer}}
& \multicolumn{3}{c}{\textbf{P3 domain}} \\
\cmidrule(lr){2-4}
\cmidrule(lr){5-7}
\cmidrule(lr){8-10}
& \scriptsize\textbf{AUROC} & \scriptsize\textbf{TPR} & \scriptsize\textbf{Cert.}
& \scriptsize\textbf{AUROC} & \scriptsize\textbf{TPR} & \scriptsize\textbf{Cert.}
& \scriptsize\textbf{AUROC} & \scriptsize\textbf{TPR} & \scriptsize\textbf{Cert.} \\
\midrule

LogReg
& \bmetric{1.00}{0.00} & \bmetric{1.00}{0.00} & \bmetric{1.00}{0.00}
& \metric{0.75}{0.19} & \bmetric{0.40}{0.26} & \bmetric{0.85}{0.25}
& \metric{0.98}{0.01} & \bmetric{0.94}{0.03} & \bmetric{1.00}{0.00} \\

Maha
& \bmetric{1.00}{0.00} & \bmetric{1.00}{0.00} & \bmetric{1.00}{0.00}
& \bmetric{0.84}{0.02} & \metric{0.00}{0.00} & \metric{0.00}{0.00}
& \bmetric{0.99}{0.01} & \metric{0.92}{0.09} & \bmetric{1.00}{0.00} \\

PCA
& \bmetric{1.00}{0.00} & \bmetric{1.00}{0.00} & \bmetric{1.00}{0.00}
& \metric{0.81}{0.04} & \metric{0.00}{0.00} & \metric{0.00}{0.00}
& \bmetric{0.99}{0.01} & \metric{0.89}{0.14} & \bmetric{1.00}{0.00} \\

KNN
& \metric{0.99}{0.01} & \metric{0.98}{0.04} & \bmetric{1.00}{0.00}
& \metric{0.80}{0.05} & \metric{0.00}{0.00} & \metric{0.00}{0.00}
& \metric{0.94}{0.05} & \metric{0.67}{0.19} & \metric{0.85}{0.14} \\

MC-drop
& \metric{0.83}{0.24} & \metric{0.81}{0.26} & \metric{0.80}{0.27}
& \metric{0.74}{0.18} & \metric{0.32}{0.25} & \metric{0.76}{0.41}
& \metric{0.78}{0.04} & \metric{0.37}{0.09} & \bmetric{1.00}{0.00} \\

\bottomrule
\end{tabular}

\vspace{0.6em}

\begin{tabular}{@{}lccccccccc@{}}
\toprule
\multirow{2}{*}{\textbf{Readout}}
& \multicolumn{3}{c}{\textbf{P4 intent}}
& \multicolumn{3}{c}{\textbf{P5 paraphrase}}
& \multicolumn{3}{c}{\textbf{P6 null}} \\
\cmidrule(lr){2-4}
\cmidrule(lr){5-7}
\cmidrule(lr){8-10}
& \scriptsize\textbf{AUROC} & \scriptsize\textbf{TPR} & \scriptsize\textbf{Cert.}
& \scriptsize\textbf{AUROC} & \scriptsize\textbf{TPR} & \scriptsize\textbf{Cert.}
& \scriptsize\textbf{AUROC} & \scriptsize\textbf{TPR} & \scriptsize\textbf{Cert.} \\
\midrule

LogReg
& \metric{0.45}{0.01} & \metric{0.04}{0.01} & \metric{0.01}{0.01}
& \metric{0.60}{0.04} & \metric{0.13}{0.04} & \metric{0.40}{0.17}
& \bmetric{0.50}{0.01} & \bmetric{0.06}{0.01} & \bmetric{0.02}{0.02} \\

Maha
& \bmetric{0.70}{0.01} & \bmetric{0.18}{0.04} & \bmetric{0.76}{0.11}
& \bmetric{1.00}{0.00} & \bmetric{1.00}{0.00} & \bmetric{1.00}{0.00}
& \bmetric{0.50}{0.01} & \metric{0.05}{0.01} & \metric{0.01}{0.01} \\

PCA
& \metric{0.62}{0.01} & \metric{0.12}{0.03} & \metric{0.41}{0.33}
& \metric{0.98}{0.01} & \metric{0.89}{0.04} & \bmetric{1.00}{0.00}
& \bmetric{0.50}{0.01} & \metric{0.05}{0.01} & \metric{0.00}{0.01} \\

KNN
& \metric{0.60}{0.01} & \metric{0.09}{0.02} & \metric{0.20}{0.22}
& \metric{0.95}{0.01} & \metric{0.73}{0.10} & \bmetric{1.00}{0.00}
& \bmetric{0.50}{0.01} & \metric{0.05}{0.01} & \metric{0.01}{0.01} \\

MC-drop
& \metric{0.46}{0.01} & \metric{0.04}{0.01} & \metric{0.01}{0.01}
& \metric{0.58}{0.04} & \metric{0.14}{0.04} & \metric{0.43}{0.23}
& \metric{0.49}{0.04} & \bmetric{0.06}{0.01} & \metric{0.01}{0.02} \\

\bottomrule

\end{tabular}
\caption{\textbf{Selected-layer readout geometry.}
All readouts use the same selected hidden layer and the same conformal/e-process pipeline. 
Each cell reports AUROC / OOD-TPR@$\tau$ / certificate rate as mean$\pm$seed-std.
Larger values indicate stronger rejection evidence for P1--P4. For P5, high rejection reflects sensitivity to rewrite-induced distribution shift; for P6, chance-level AUROC and low certification are expected.}
\label{tab:selected_layer_readouts}
\end{table*}

\subsection{Geometry of Selected-Layer OOD Signals}
\label{sec:ood_signal_geometry}

This section answers \textbf{RQ3}: why different boundary definitions lead to different rejection behavior. 
The cross-model audit uses CLG as the primary gate because it gives the strongest controlled operating point in Sec.~\ref{sec:controlled_baseline}. 
Here, additional readouts are used for analysis rather than post-hoc model selection: we keep the selected layer fixed and vary only the score family, so that each readout probes a different geometry of the same hidden representation.

The readouts expose complementary signal types. 
LogReg tests whether the held-out boundary aligns with the linear OOD direction learned from the representative development OOD source.
Mahalanobis tests whether OOD examples move away from the IND covariance ellipsoid. 
PCA tests whether they leave the low-dimensional manifold occupied by IND representations. 
KNN tests whether they fall outside the local neighborhood structure of IND examples. 
The three metrics capture different stages of the gate: AUROC measures offline ranking, OOD-TPR@$\tau$ measures rejection under the IND-calibrated conformal threshold, and certificate rate measures whether calibrated rejections persist strongly enough to form sequential evidence.

\begin{figure}[t]
    \centering
    \includegraphics[width=\columnwidth]{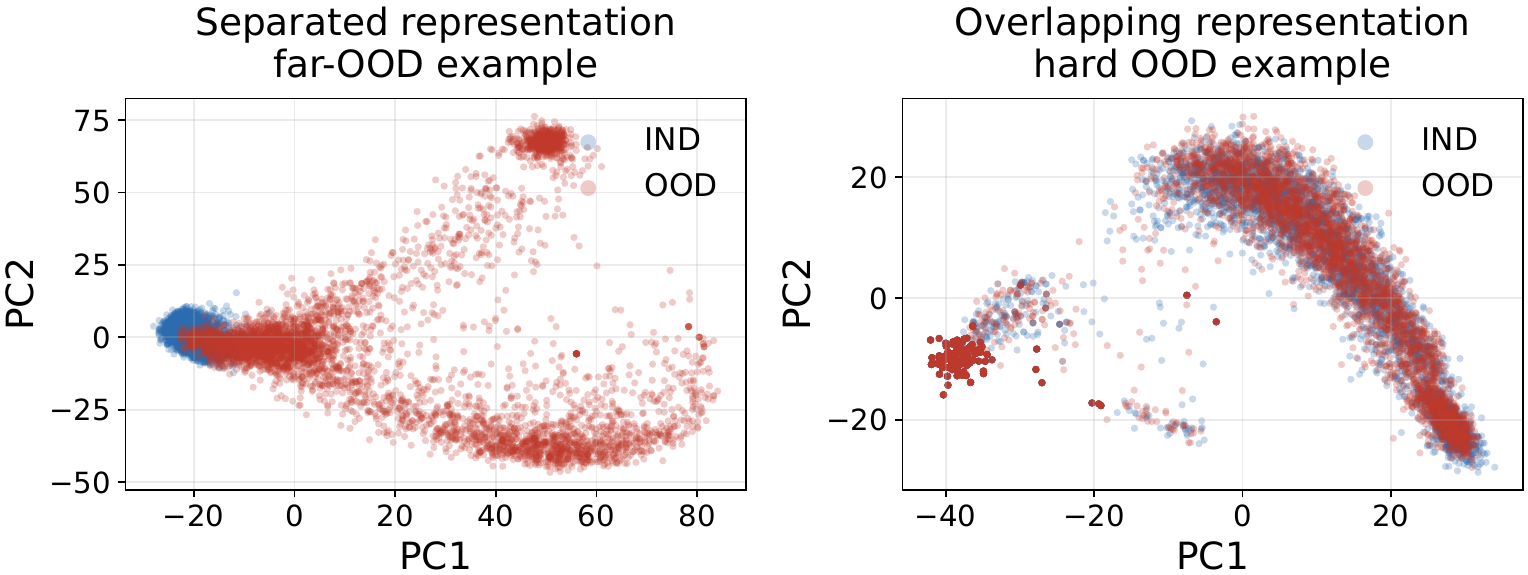}
\caption{\textbf{Selected-layer representation geometry.}
PCA projections illustrate separable and overlapping IND/OOD hidden-state patterns at the selected layer.}
    \label{fig:pca_geometry_main}
\end{figure}

Figure~\ref{fig:pca_geometry_main} gives a qualitative view. 
Broad OOD settings show clearer IND/OOD separation at the selected layer, while tight intent-level settings remain heavily overlapping. 
This visual pattern matches the certification gap observed in Sec.~\ref{sec:cross_model_rejection}.

Table~\ref{tab:selected_layer_readouts} separates the boundary effects into three regimes. 
First, P1 and P3 are representation-strong regimes: several readouts achieve high AUROC, high OOD-TPR@$\tau$, and high certificate rate, showing that the selected layer contains a stable OOD signal that is not tied to a particular score family.

Second, P2 shows why calibrated rejection is stricter than offline ranking. 
Mahalanobis, PCA, and KNN obtain reasonable AUROC, but their OOD-TPR@$\tau$ and certificate rates collapse under the IND-calibrated threshold. 
LogReg and MC-dropout retain stronger sequential evidence because their scores more often cross the conformal threshold on the held-out MNLI stream. 
Thus, transfer to a new OOD source depends not only on ranking quality, but also on persistent calibrated rejection.

Third, P4 and P5 explain why boundary definition matters. 
P4 is a fine-grained Banking77 intent split. 
Distance-based readouts recover some signal, with Mahalanobis reaching 0.70 AUROC and 0.76 certificate rate, but OOD-TPR@$\tau$ remains much lower than in P1 and P3. 
This indicates a tight intent boundary: the selected representation contains partial distance-based evidence, but high-recall rejection under an IND-calibrated threshold is harder than for broader domain shifts.

P5 has a different geometry. 
The paraphrases preserve the original Banking77 intent but change surface form. 
Mahalanobis, PCA, and KNN reveal a strong rewrite-induced distribution shift, with near-perfect AUROC and certificate rate. 
This means the selected layer encodes the paraphrase shift, but the signal is better read as distance, manifold, or neighborhood deviation than as the linear OOD readout learned from the representative development OOD source.
The weaker LogReg rejection on P5 is therefore different from the weak rejection on P4: P4 changes the intent boundary itself, whereas P5 preserves the service intent while altering the input realization. 
Thus, P5 is not simply a harder version of P4; it decouples intent-level service support from representation-level distribution shift.

Finally, P6 confirms that changing the readout does not create artificial evidence on same-distribution traffic. 
All score families remain near chance AUROC with near-zero certification.

Overall, this analysis shows that SCOPE does more than report a single OOD score: it shows which service boundaries produce readable selected-layer evidence and what geometric form that evidence takes.

\section{Conclusion}

In this paper, we argue that OOD rejection for LLM services should be treated not merely as an offline scoring problem, but as a calibrated service-boundary gating problem. 
Across multiple LLMs and boundary conditions, \textsc{SCOPE} improves gate-level reliability over standard final-output and final-layer detectors, and reveals how different OOD boundaries are encoded in hidden space.
Beyond rejection performance, this representation-level view provides a lightweight interpretability lens for understanding where service-boundary information appears inside frozen LLMs.
More broadly, \textsc{SCOPE} offers a statistical layer for building LLM services that can reject, route, or escalate unsupported traffic before full generation, and can be extended to multimodal and tool-augmented systems under non-stationary real-world traffic.

\newpage

\section*{Limitations}

SCOPE is representation-relative: it certifies persistent rejection evidence for a fixed frozen model, selected layer, score rule, and conformal threshold, rather than proving that an IND/OOD boundary is intrinsically separable in input space. 
Its power depends on whether the selected representation and readout expose the relevant service-boundary signal, and on how representative the development OOD source is for the held-out boundary.

The statistical guarantees also rely on standard calibration and sequential-testing assumptions. 
The conformal threshold controls IND false rejection under exchangeability between calibration and future IND traffic, while the e-process is valid for a fixed gate and fixed \(p_1\) under the operational null that rejection probability is bounded by the IND-calibrated baseline. 
Our experiments use offline held-out streams with shuffled arrival orders; real service traffic may be non-stationary, adaptive, or affected by routing feedback. 
Future work should study SCOPE in online deployments, multimodal services, and tool-augmented agent systems.

\section*{Acknowledgments}

This work is partially funded by the European Union (under grant agreement ID 101212818). Views and opinions expressed are however those of the author(s) only and do not necessarily reflect those of the European Union or European Health and Digital Executive Agency (HADEA). Neither the European Union nor the granting authority can be held responsible for them. This work is partially supported by Innovate UK through AI-PASSPORT under Grant 10126404. This work was awarded a grant by the AI Security Institute (AISI) via the Alignment Project (Rare-Event Estimation in Large Language Models via Subset Simulation) and funded by EPSRC. Yi's contribution is partially supported through the Royal Society international exchanges programme and in part by the Engineering and Physical Sciences Research Council, through funding from RAi UK [EP/Y009800/1].

\bibliography{custom}

\appendix

\newpage
\appendix

\section{Scaling Across Model Sizes}
\label{app:model_scaling}

We further examine whether the selected-layer behavior changes systematically with model size. This analysis is complementary to the cross-model audit in the main text. We repeat the CLG evaluation on the \textsc{Qwen2.5} instruction-tuned family with 1.5B, 7B, and 14B parameters. 
For each model, we evaluate three representative IND/OOD pairs: a far-OOD pair (SST-2$\rightarrow$TREC), a near-intent pair (Banking77$\rightarrow$Banking77-OOD), and a cross-task pair (20 Newsgroups$\rightarrow$MNLI). 
For each pair and model, we select the best layer $\ell^\star$ using development AUROC and then record the CLG AUROC, the certification stopping time $T_\alpha$, and the relative selected depth $\ell^\star/L$.

\begin{figure}[t]
    \centering
    \includegraphics[width=\columnwidth]{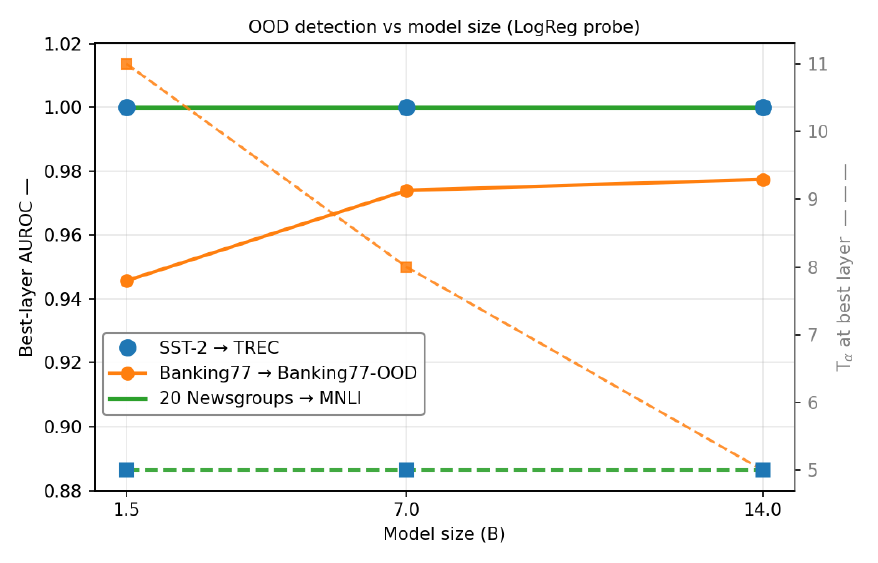}
    \includegraphics[width=0.9\columnwidth]{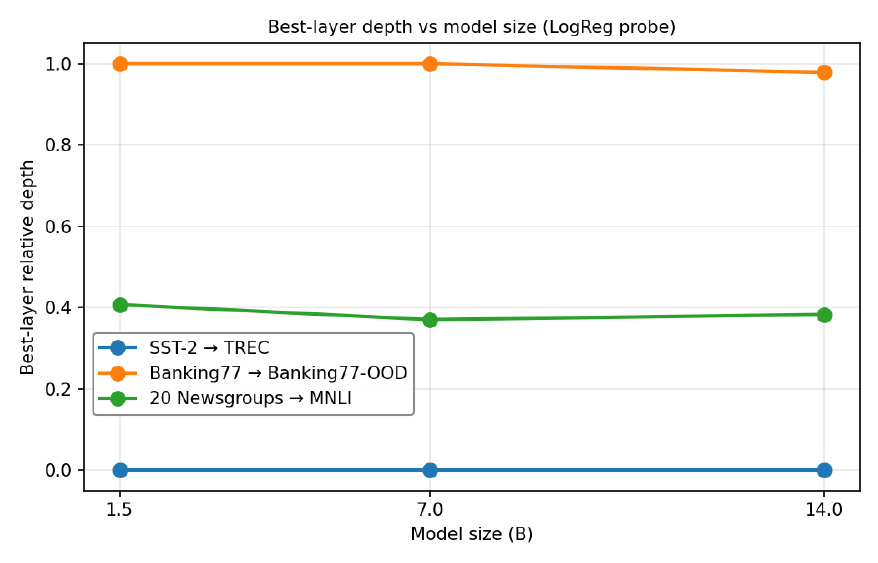}
    \caption{\textbf{Scaling behavior of CLG across the \textsc{Qwen2.5} family.}
    Left: best-layer AUROC and certification stopping time $T_\alpha$ across model sizes for representative IND/OOD pairs. 
    Right: relative depth of the selected layer $\ell^\star/L$. 
    Larger models generally produce stronger OOD separation and faster certification, while the relative location of the selected layer remains qualitatively stable within the model family.}
    \label{fig:qwen_scaling}
\end{figure}

Figure~\ref{fig:qwen_scaling} provides two useful checks. 
First, AUROC generally improves and $T_\alpha$ decreases as model size increases, suggesting that larger backbones tend to encode more readable service-boundary signals. 
Second, the relative depth of the selected layer remains qualitatively stable across the \textsc{Qwen2.5} family for the same boundary type. 
This supports the use of a one-time layer-selection step and reinforces the main claim that OOD evidence is representation-dependent rather than necessarily concentrated at the final layer.

\section{Component Ablations}

\subsection{Effect of conformal calibration}
\label{app:tau_sensitivity}

We next study how the conformal false-rejection budget $\epsilon$ affects the operating point of the gate. 
The underlying CLG scores are fixed; only the conformal threshold $\tau$ changes with $\epsilon$. 
Thus, AUROC remains unchanged, while OOD-TPR@$\tau$ and certificate rate vary because the rejection rule becomes stricter or looser.

\begin{figure*}[t]
    \centering
    \includegraphics[width=\textwidth]{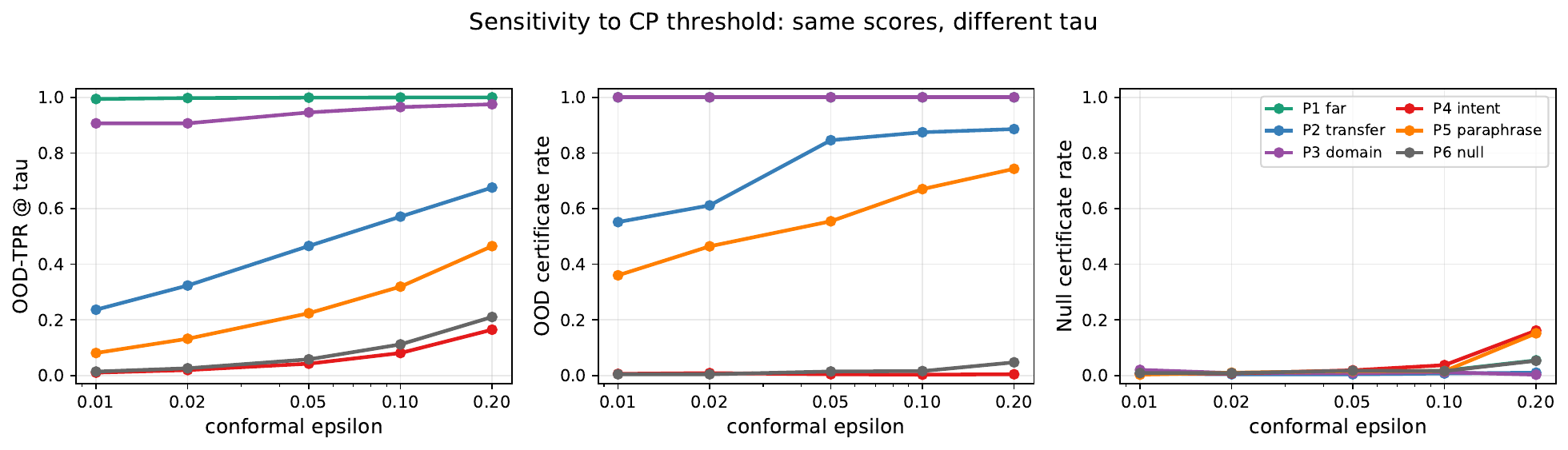}
    \caption{\textbf{Sensitivity to the conformal threshold.}
    Smaller $\epsilon$ gives a stricter threshold, reducing IND false rejections but also lowering OOD-TPR and certificate rate. 
    Larger $\epsilon$ increases OOD rejection and certification by spending more IND false-rejection budget. 
    The main experiments use $\epsilon=0.05$ as a balanced operating point.}
    \label{fig:tau_sensitivity}
\end{figure*}

Figure~\ref{fig:tau_sensitivity} shows the expected calibration trade-off. 
A smaller $\epsilon$ makes the gate more conservative: it protects IND traffic more strongly but rejects fewer OOD examples, which slows or prevents e-process certification. 
A larger $\epsilon$ makes the gate more permissive and increases OOD certificates, but it uses more of the allowed IND false-rejection budget. 
This analysis strengthens the deployment interpretation of SCOPE: $\epsilon$ is not a post-hoc ROC threshold, but an explicit operating budget chosen before stream evaluation.

\begin{figure}[t]
\centering
\includegraphics[width=\linewidth]{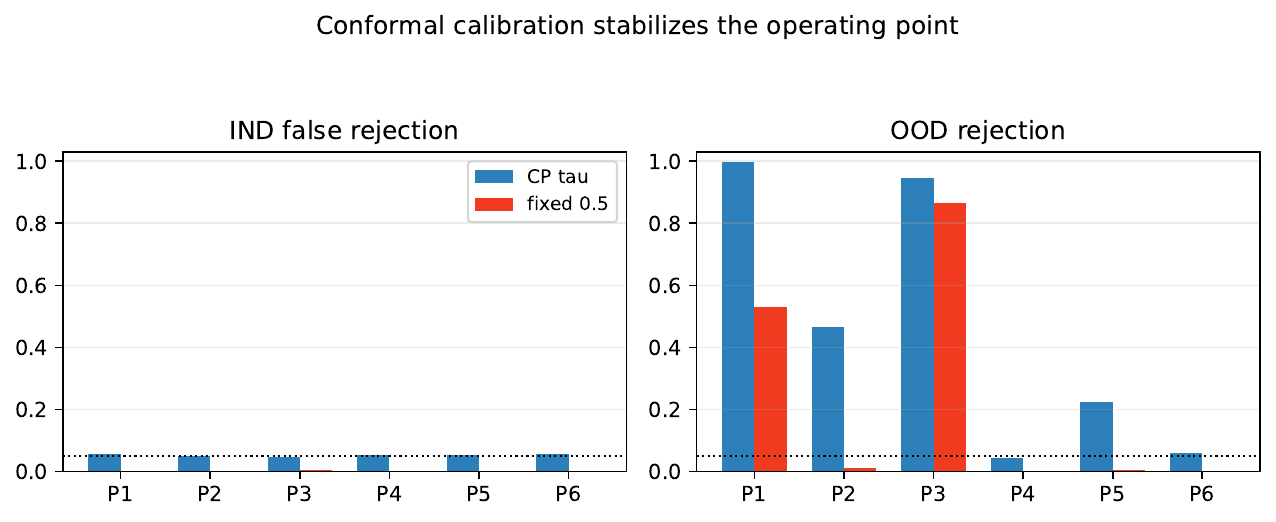}
\caption{Conformal calibration stabilizes the operating point.  A fixed raw
threshold can be overly conservative because probe score scales vary across
pairs and models.  The conformal threshold adapts the same score family to a
target IND false rejection budget.}
\label{fig:ablation_cp_vs_fixed}
\end{figure}

\input{Tab/CP_ablation}

Table~\ref{tab:cp-roc-comparison} evaluates the calibration layer of the gate. 
A high AUROC score is useful only if it can be converted into an operating point with controlled false rejections on IND inputs. 
The conformal threshold keeps empirical IND false rejection close to the target budget while maintaining high OOD-TPR across the evaluated settings. 
By contrast, ROC-tuned thresholds may look attractive on a development split, but they do not provide finite-sample IND-side control. 
In several cases, they reduce IND-FPR by sacrificing OOD recall.

This distinction is central to SCOPE. 
The goal is not only to find a threshold that works on one validation set, but to obtain a rejection rule whose IND-side error is explicitly calibrated before held-out stream evaluation. 
The e-process then operates on this fixed conformalized rejection rule.

\subsection{Sensitivity to $p_1$, and $\alpha$}

The e-process parameters $p_1$ and $\alpha$ control the evidence accumulation
after $\tau$ is fixed.  The parameter $p_1$ is the alternative rejection-rate
bet: smaller values are more sensitive to weak but persistent deviations, while
larger values reward stronger rejection streams and can cross faster when the
shift is obvious.  The significance level $\alpha$ sets the evidence threshold
$1/\alpha$; larger $\alpha$ lowers the crossing threshold and increases the
certificate rate, while smaller $\alpha$ asks for stronger evidence.  Across
these sweeps, P1 and P3 remain stable, P4 remains difficult, and P6 stays low
under the default setting.  We therefore use $\epsilon=0.05$, $p_1=0.30$, and
$\alpha=0.01$ as a balanced default in the main experiments.

\begin{figure}[t]
\centering
\includegraphics[width=\linewidth]{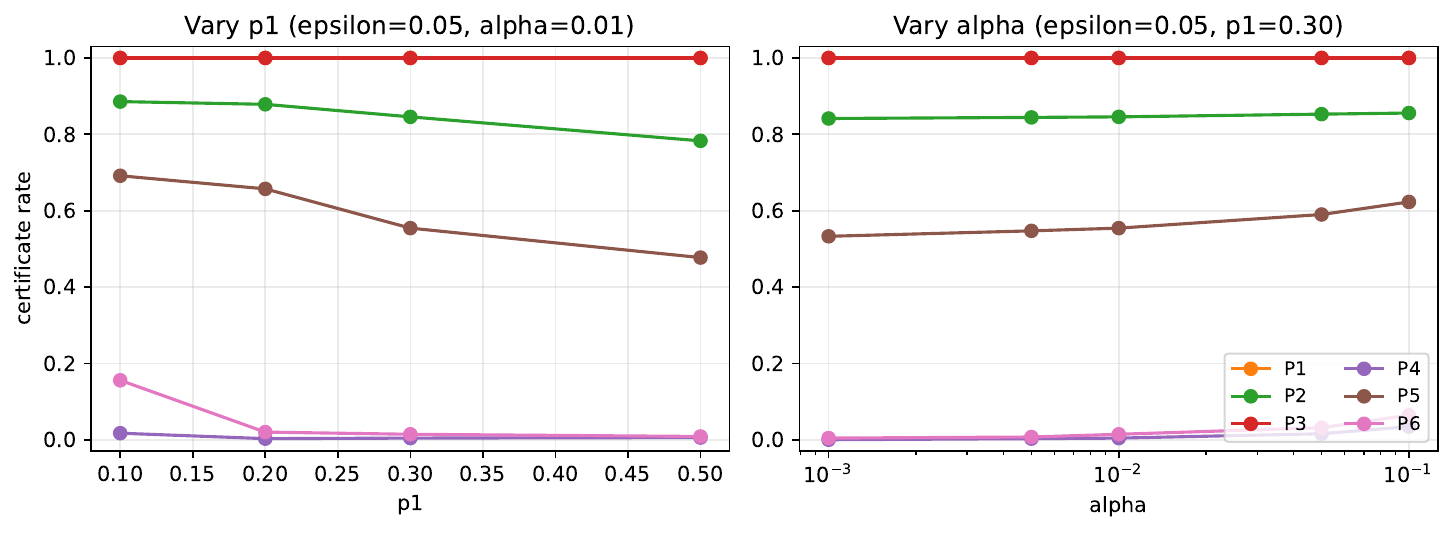}
\caption{Sensitivity to the e-process parameters.  Varying $p_1$ changes the
alternative rejection-rate bet, and varying $\alpha$ changes the evidence
threshold $1/\alpha$.  The default $(p_1,\alpha)=(0.30,0.01)$ gives a stable
balance between sensitivity and null safety.}
\label{fig:ablation_p1_alpha_sensitivity}
\end{figure}

\section{Additional Readout and Geometry Analysis}

\subsection{Additional readout analysis for P4 and P5}
\label{app:add_readout_p4p5}
\begin{figure}[h]
    \centering
    \includegraphics[width=\columnwidth]{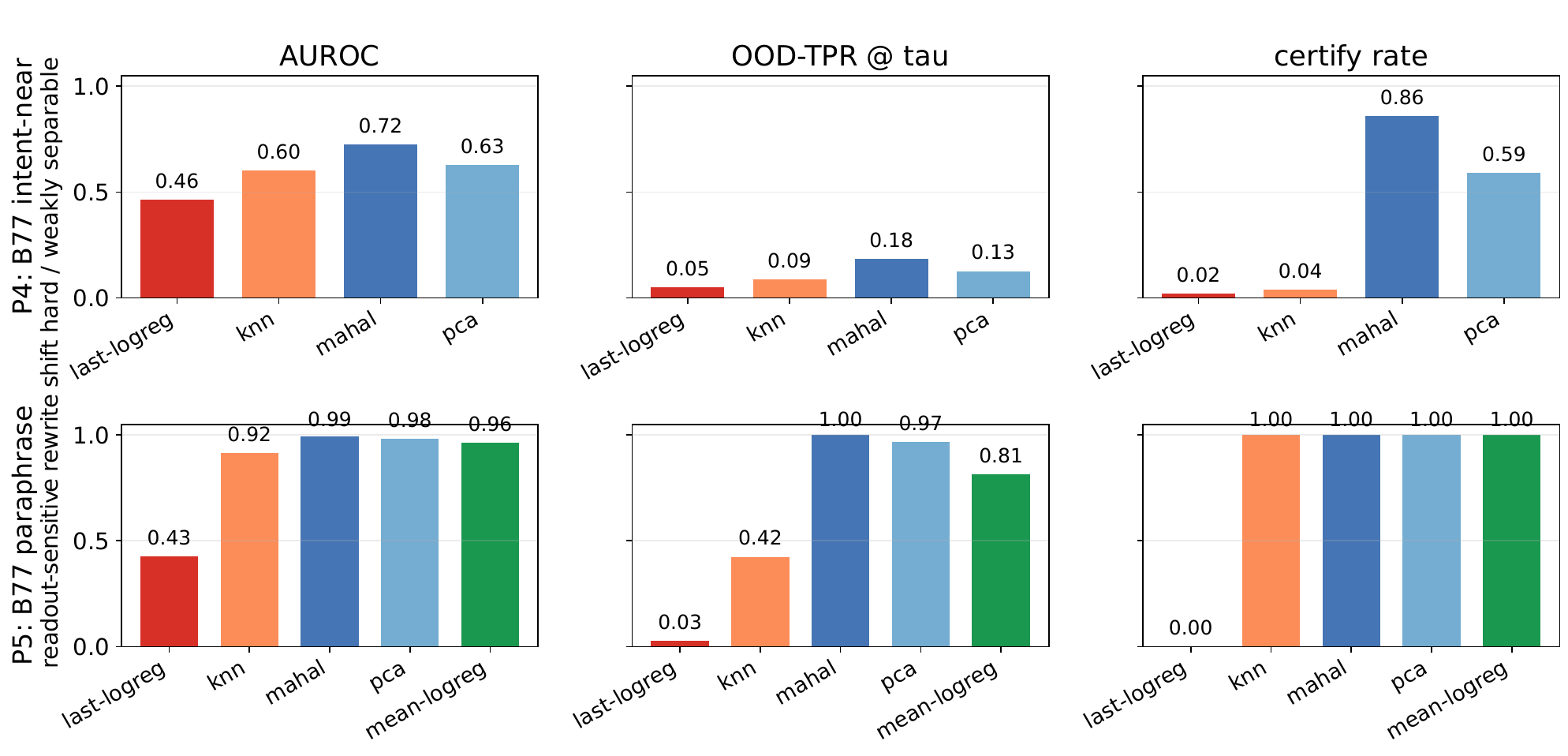}
    \caption{\textbf{Additional readout analysis for P4 and P5.}
    P4 remains a fine-grained intent boundary: alternative readouts improve ranking and certification only partially, and calibrated OOD-TPR remains limited. 
    P5 shows a different pattern: paraphrase-induced shifts are strongly visible to distance-based, manifold-based, and pooled readouts, indicating that the selected layer encodes rewrite-induced distribution shift even when the default last-token linear gate is less sensitive.}
    \label{fig:p4p5_readout}
\end{figure}

Figure~\ref{fig:p4p5_readout} gives a closer view of the two boundary-sensitive cases discussed in Sec.~\ref{sec:ood_signal_geometry}. 
All variants use the same selected layer and the same conformal/e-process protocol. 
The only change is the readout family, which lets us examine what type of hidden-state shift is present.

For P4, the Banking77 intent-near split, changing the score family improves offline ranking only moderately and yields limited OOD-TPR at the conformal threshold. 
This supports the interpretation that P4 is a genuinely fine-grained service boundary: the selected representation contains some distance-based evidence, but IND and OOD intents remain closely entangled under an IND-calibrated operating point.

For P5, the paraphrase stress test, the behavior is different. 
The paraphrased inputs preserve the original intent but introduce a systematic rewrite-induced shift. 
Distance-based readouts and mean-pooling variants recover strong evidence, showing that the selected layer encodes this shift even when the default last-token linear readout is less sensitive to it. 
Thus, P5 is best interpreted as a boundary-definition and readout-geometry case rather than a standard unsupported-intent OOD benchmark.

The comparison between P4 and P5 shows that similar LogReg behavior can arise from different boundary semantics. 
For P4, weak LogReg rejection reflects the fine-grained nature of the Banking77 intent split. 
For P5, the paraphrases preserve the original service intent, so weak LogReg rejection is consistent with the gate not treating intent-preserving rewrites as unsupported intents. 
The strong Mahalanobis, PCA, and KNN results nevertheless show that the selected layer encodes a rewrite-induced distribution shift. 
This supports our view that service-boundary definitions determine not only detection difficulty, but also which hidden-state geometry is relevant.

\subsection{Additional PCA geometry}
\label{app:pca_examples}

We include additional PCA projections to visualize how selected-layer representations differ across easy and tight OOD boundaries. 
Each plot shows the first two principal components of last-token embeddings at the selected layer.

\begin{figure}[h]
  \centering
  \begin{subfigure}[b]{0.495\columnwidth}
    \centering
    \includegraphics[width=\linewidth]{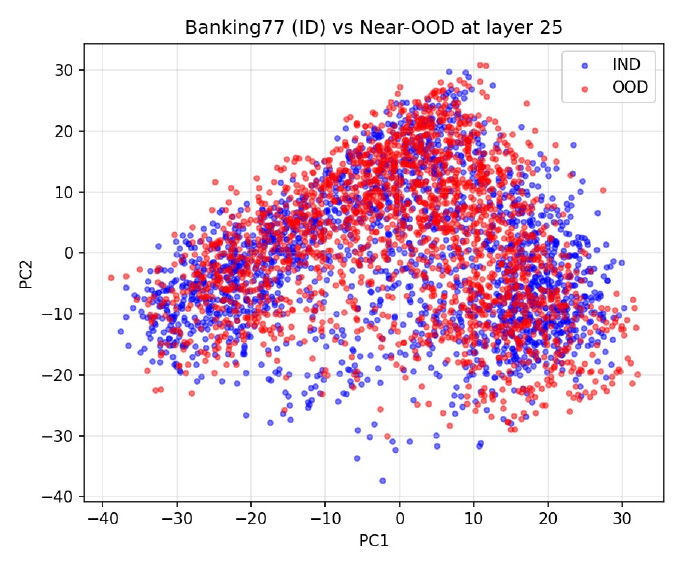}
  \end{subfigure}
  \hfill
  \begin{subfigure}[b]{0.475\columnwidth}
    \centering
    \includegraphics[width=\linewidth]{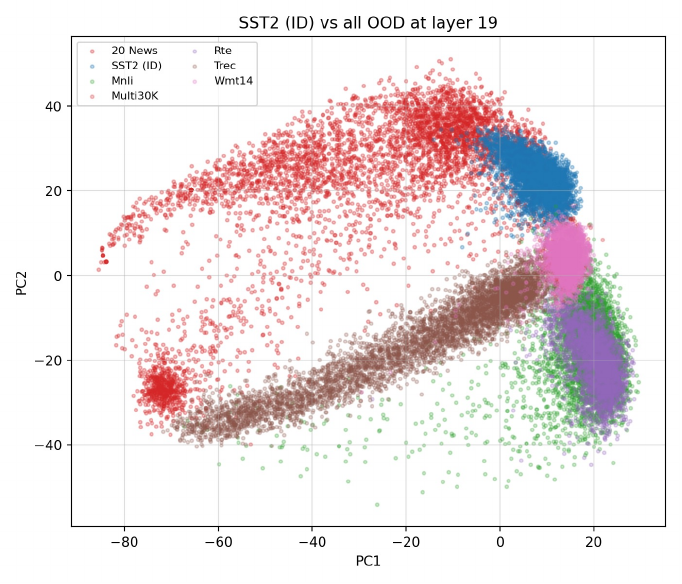}

  \end{subfigure}
    \caption{\textbf{Additional PCA examples of selected-layer IND/OOD geometry.}
    Broad OOD boundaries show clearer separation in the selected representation, while tight near-intent boundaries remain more entangled. 
    These examples complement the main geometry analysis by illustrating why calibrated rejection is easier for broad service shifts than for fine-grained intent splits.}
    \label{fig:pca_examples}
\end{figure}

The PCA views are not used for model selection or threshold tuning. 
They serve only as qualitative support for the readout analysis in Sec.~\ref{sec:ood_signal_geometry}. 
When IND and OOD examples are visibly separated, multiple readouts tend to produce high OOD-TPR and strong certificates. 
When the two groups remain entangled, high-recall rejection under an IND-calibrated threshold becomes more demanding.

\subsection{Directional gate e-process}
\label{app:directional_eprocess}

We also examine the e-process behavior of the low-capacity directional variant. 
The directional gate reads a one-dimensional OOD direction from the selected representation, while CLG uses a full linear boundary. 
This comparison illustrates why readout capacity matters even when the layer is fixed.

\begin{figure}[h]
    \centering
    \includegraphics[width=\columnwidth]{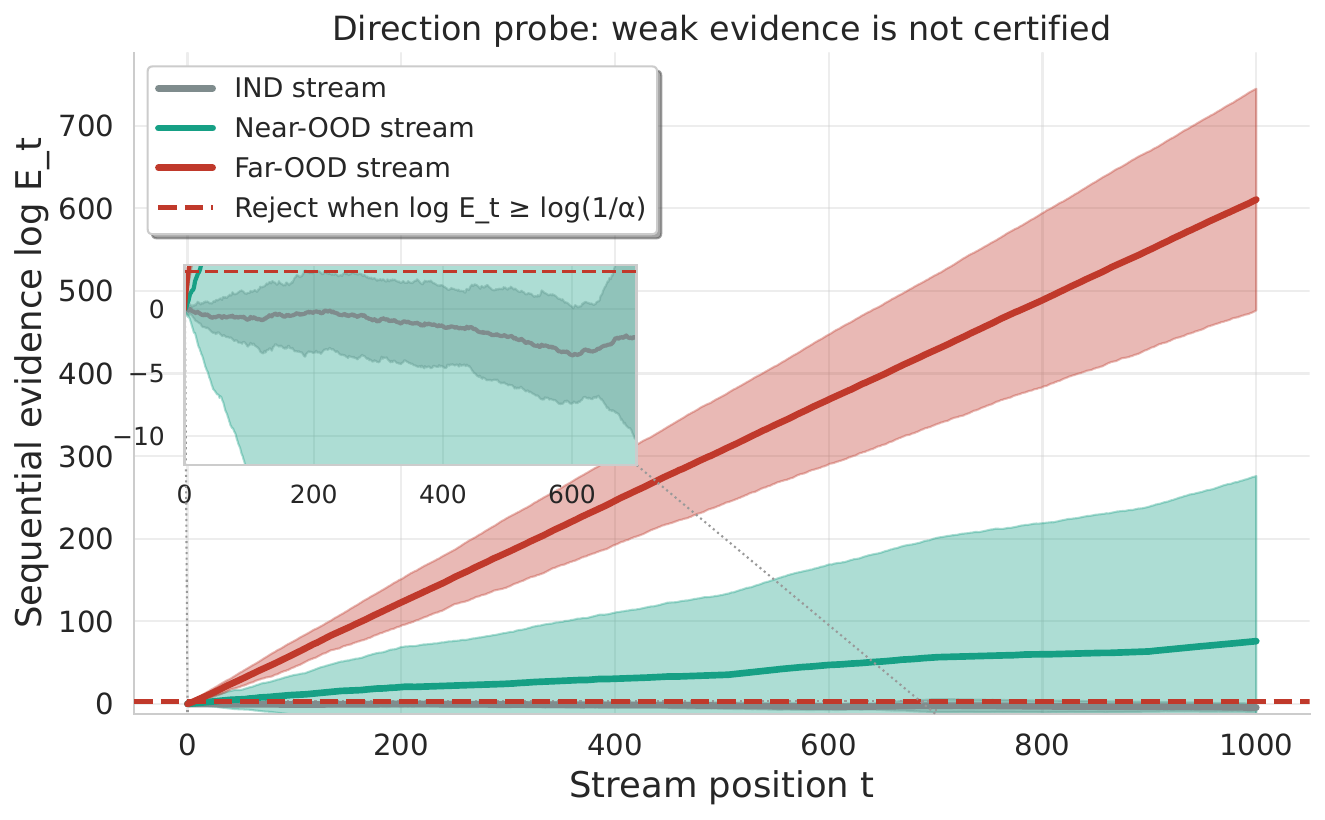}
    \caption{\textbf{E-process trace for the conformal directional gate.}
    The directional gate accumulates strong evidence on broad far-OOD streams, but produces weaker evidence on tighter near-OOD streams. 
    This supports the use of CLG as the main gate: a one-dimensional direction can capture coarse shifts, while a richer linear boundary is more effective for finer service-boundary signals.}
    \label{fig:directional_eprocess}
\end{figure}

Figure~\ref{fig:directional_eprocess} shows that the directional variant can certify broad OOD shifts but is less effective on tighter boundaries. 
This is consistent with the main controlled comparison: selected layers often contain OOD-sensitive directions, but fine-grained boundaries require a richer readout. 
The result strengthens the motivation for CLG as the default instantiation of SCOPE.

\begin{table*}[t]
\centering
\small
\setlength{\tabcolsep}{4.5pt}
\renewcommand{\arraystretch}{1.12}
\resizebox{\textwidth}{!}{%
\begin{tabular}{lllll}
\toprule
Boundary & IND domain & Dev OOD source & Held-out stream & Boundary type \\
\midrule
P1 & SST-2 & RTE & WMT14 De-En & far-OOD transfer \\
P2 & 20 Newsgroups & TREC & MNLI & cross-task OOD transfer \\
P3 & CLINC150 Travel & CLINC150 near split & CLINC150 Banking & near-domain intent shift \\
P4 & Banking77, 38 IND intents & Banking77 OOD-side training split & Banking77, 39 held-out intents & fine-grained intent boundary \\
P5 & Banking77, same IND side as P4 & 20 Newsgroups & 500 Banking77 paraphrases & intent-preserving rewrite shift \\
P6 & SST-2 first 5k examples & RTE & SST-2 held-out second 5k examples & same-distribution null stream \\
\bottomrule
\end{tabular}
}
\caption{\textbf{Service-boundary conditions used in the evaluation.}
The development OOD source is used for readout fitting and layer selection, while the held-out stream is used only for evaluation.}
\label{tab:boundary_conditions}
\end{table*}

\section{Service-Boundary Construction}
\label{app:boundary_construction}

Table~\ref{tab:boundary_conditions} summarizes the P1--P6 service-boundary conditions used in Secs.~\ref{sec:cross_model_rejection} and~\ref{sec:ood_signal_geometry}. These settings are described in Sec.~\ref{sec:exp_setup}, but we repeat them here to make the paper easier to follow.

\section{Compute resources}
\label{app:compute_resources}
All experiments are conducted in Python\textsuperscript{\textregistered} on a machine equipped with an AMD EPYC\textsuperscript{\textregistered} 7452 32-Core Processor, 128GB of RAM, and one A100 GPU with 40GB of VRAM.

\section{Deployment Overhead}
\label{app:deployment_overhead}

To assess deployment efficiency, we measure the online latency overhead introduced by the selected-layer gate relative to the base model forward pass. 
All methods are evaluated on the same hardware with the same batch size and input length. 
The one-time cost of layer selection is excluded from online latency because it is performed before deployment and amortized across future requests.

\input{Tab/overhead}

Table~\ref{tab:deployment_overhead} shows that CLG adds only a small online overhead over the base forward pass. 
Its latency is comparable to simple representation-based scores such as cosine distance and lower than Mahalanobis scoring, while providing IND-side conformal calibration and e-process certification. 
This supports the intended use of SCOPE as a lightweight front-end gate rather than a replacement for the underlying LLM.

\end{document}

%% file: Tab/CP_ablation.tex
\begin{table}[h]
\centering
\resizebox{0.9\columnwidth}{!}{%
\begin{tabular}{@{}lcccc@{}}
\toprule
\multirow{2}{*}{IND Dataset} & \multicolumn{2}{c}{Conformal} & \multicolumn{2}{c}{ROC-tuned} \\ \cmidrule(lr){2-3} \cmidrule(lr){4-5}
 & IND\_FPR & OOD\_TPR & IND\_FPR & OOD\_TPR \\ \midrule
20NG & 0.0450 & \textbf{1.0000} & \textbf{0.0000} & 0.9340 \\
SST2 & 0.0400 & \textbf{1.0000} & \textbf{0.0000} & 0.9540 \\
Clinc-Banking & 0.0857 & \textbf{1.0000} & \textbf{0.0000} & 0.9429 \\
Clinc-Travel & 0.0111 & \textbf{1.0000} & \textbf{0.0000} & 0.9407 \\
Banking77 & \textbf{0.0142} & 0.9446 & \textbf{0.0142} & \textbf{0.9486} \\ \bottomrule
\end{tabular}
}
\caption{\textbf{Conformal calibration versus ROC-tuned thresholding.}
The conformal threshold keeps IND false rejections close to the target level while preserving high OOD recall. ROC-tuned thresholds can reduce IND-FPR on the validation split, but they do not provide finite-sample IND-side control and may sacrifice OOD-TPR.}
\label{tab:cp-roc-comparison}
\label{tab:cp_ablation}
\end{table}

%% file: Tab/overhead.tex
\begin{table}[h]
\centering
\resizebox{0.95\columnwidth}{!}{%
\begin{tabular}{lccc}
\toprule
Method & Latency & Overhead & Offline Cost \\
\midrule
Base model only & 9.6 ms & 0.0 \% & None \\
MSP & 9.7 ms & 0.6 \% & None \\
Energy & 9.7 ms & 0.9 \% & None \\
Mahalanobis & 10.0 ms & 3.6 \% & Precomputed stats \\
Cosine & 9.8 ms & 1.4 \% & Precomputed stats \\
\textbf{CLG} & 9.8 ms & 1.5 \% & One-time layer search \\
\bottomrule
\end{tabular}
}
\caption{Deployment-time efficiency comparison.}
\label{tab:deployment_overhead}
\end{table}

%% file: custom.bib
@inproceedings{wang2023hallucination,
    title = "Hallucination Detection for Generative Large Language Models by {B}ayesian Sequential Estimation",
    author = "Wang, Xiaohua  and
      Yan, Yuliang  and
      Huang, Longtao  and
      Zheng, Xiaoqing  and
      Huang, Xuanjing",
    editor = "Bouamor, Houda  and
      Pino, Juan  and
      Bali, Kalika",
    booktitle = "Proceedings of the 2023 Conference on Empirical Methods in Natural Language Processing",
    month = dec,
    year = "2023",
    address = "Singapore",
    publisher = "Association for Computational Linguistics",
    url = "https://aclanthology.org/2023.emnlp-main.949/",
    doi = "10.18653/v1/2023.emnlp-main.949",
    pages = "15361--15371"
}

@inproceedings{dai2023exploring,
	author       = {Yi Dai and Hao Lang and Kaisheng Zeng and Fei Huang and Yongbin Li},
	title        = {Exploring Large Language Models for Multi-Modal Out-of-Distribution Detection},
	booktitle    = {Findings of the Association for Computational Linguistics: EMNLP 2023},
	year         = 2023,
	pages        = {5292--5305}
}

@inproceedings{li2024referencefree,
	author       = {Qing Li and Jiahui Geng and Chenyang Lyu and Derui Zhu and Maxim Panov and Fakhri Karray},
	title        = {Reference-free Hallucination Detection for Large Vision-Language Models},
	booktitle    = {Findings of the Association for Computational Linguistics: EMNLP 2024},
	year         = 2024,
	pages        = {4542--4551}
}

@inproceedings{deng2024dontknow,
	author       = {Yang Deng and Yong Zhao and Moxin Li and See{-}Kiong Ng and Tat{-}Seng Chua},
	title        = {Don't Just Say ``\text{I} Don't Know''! Self-Aligning Large Language Models for Responding to Unknown Questions with Explanations},
	booktitle    = {Proceedings of the 2024 Conference on Empirical Methods in Natural Language Processing},
	year         = 2024
}

@inproceedings{sali2025navigating,
    title = "Navigating the Unknown: Intent Classification and Out-of-Distribution Detection Using Large Language Models",
    author = "Sali, Yusuf  and
      Toraman, S{\i}tk{\i} Can",
    editor = "Christodoulopoulos, Christos  and
      Chakraborty, Tanmoy  and
      Rose, Carolyn  and
      Peng, Violet",
    booktitle = "Findings of the Association for Computational Linguistics: EMNLP 2025",
    month = nov,
    year = "2025",
    address = "Suzhou, China",
    publisher = "Association for Computational Linguistics",
    url = "https://aclanthology.org/2025.findings-emnlp.791/",
    doi = "10.18653/v1/2025.findings-emnlp.791",
    pages = "14652--14664",
    ISBN = "979-8-89176-335-7"
}

@inproceedings{ahdritz2024knowable,
	author       = {Gustaf Ahdritz and Tian Qin and Nikhil Vyas and Boaz Barak and Benjamin L. Edelman},
	title        = {Distinguishing the Knowable from the Unknowable with Language Models},
	booktitle    = {Proceedings of the 41st International Conference on Machine Learning},
	volume       = 235,
	pages        = {503--549},
	year         = 2024
}

@inproceedings{bayat2024lito,
	author       = {Farima Fatahi Bayat and Xin Liu and H. V. Jagadish and Lu Wang},
	title        = {Enhanced Language Model Truthfulness with Learnable Intervention and Uncertainty Expression},
	booktitle    = {Findings of the Association for Computational Linguistics: ACL 2024},
	year         = 2024,
	pages        = {12388--12400}
}

@inproceedings{alain2017linearprobes,
	author       = {Guillaume Alain and Yoshua Bengio},
	title        = {Understanding Intermediate Layers Using Linear Classifier Probes},
	booktitle    = {Proceedings of the 5th International Conference on Learning Representations, Workshop Track},
	year         = 2017
}

@inproceedings{azaria2023internal,
	author       = {Amos Azaria and Tom Mitchell},
	title        = {The Internal State of an {LLM} Knows When It's Lying},
	booktitle    = {Findings of the Association for Computational Linguistics: EMNLP 2023},
	year         = 2023,
	pages        = {967--976}
}

@inproceedings{marks2023geometry,
title={The Geometry of Truth: Emergent Linear Structure in Large Language Model Representations of True/False Datasets},
author={Samuel Marks and Max Tegmark},
booktitle={First Conference on Language Modeling},
year={2024},
url={https://openreview.net/forum?id=aajyHYjjsk}
}

@inproceedings{gurnee2024spacetime,
title={Language Models Represent Space and Time},
author={Wes Gurnee and Max Tegmark},
booktitle={The Twelfth International Conference on Learning Representations},
year={2024},
url={https://openreview.net/forum?id=jE8xbmvFin}
}

@misc{bouchaud2025linear,
title={Linear socio-demographic representations emerge in Large Language Models from indirect cues}, 
      author={Paul Bouchaud and Pedro Ramaciotti},
      year={2025},
      eprint={2512.10065},
      archivePrefix={arXiv},
      primaryClass={cs.AI},
      url={https://arxiv.org/abs/2512.10065}, 
}

@inproceedings{
cencerrado2025noanswer,
title={No Answer Needed: Predicting {LLM} Answer Accuracy from Question-Only Linear Probes},
author={Iv{\'a}n Vicente Moreno Cencerrado and Arnau Padr{\'e}s Masdemont and Anton Gonzalvez Hawthorne and David Demitri Africa and Lorenzo Pacchiardi},
booktitle={ICLR 2026 Workshop on Principled Design for Trustworthy AI - Interpretability, Robustness, and Safety across Modalities},
year={2026},
url={https://openreview.net/forum?id=1QcY6LPcdQ}
}

@misc{kadavath2022know,
      title={Language Models (Mostly) Know What They Know}, 
      author={Saurav Kadavath and Tom Conerly and Amanda Askell and Tom Henighan and Dawn Drain and Ethan Perez and Nicholas Schiefer and Zac Hatfield-Dodds and Nova DasSarma and Eli Tran-Johnson and Scott Johnston and Sheer El-Showk and Andy Jones and Nelson Elhage and Tristan Hume and Anna Chen and Yuntao Bai and Sam Bowman and Stanislav Fort and Deep Ganguli and Danny Hernandez and Josh Jacobson and Jackson Kernion and Shauna Kravec and Liane Lovitt and Kamal Ndousse and Catherine Olsson and Sam Ringer and Dario Amodei and Tom Brown and Jack Clark and Nicholas Joseph and Ben Mann and Sam McCandlish and Chris Olah and Jared Kaplan},
      year={2022},
      eprint={2207.05221},
      archivePrefix={arXiv},
      primaryClass={cs.CL},
      url={https://arxiv.org/abs/2207.05221}, 
}

@inproceedings{
kuhn2023semantic,
title={Semantic Uncertainty: Linguistic Invariances for Uncertainty Estimation in Natural Language Generation},
author={Lorenz Kuhn and Yarin Gal and Sebastian Farquhar},
booktitle={The Eleventh International Conference on Learning Representations },
year={2023},
url={https://openreview.net/forum?id=VD-AYtP0dve}
}

@article{farquhar2024semantic,
	title        = {Detecting hallucinations in large language models using semantic entropy},
	author       = {Farquhar, Sebastian and Kossen, Jannik and Kuhn, Lorenz and Gal, Yarin},
	journal      = {Nature},
	volume       = 630,
	number       = 8017,
	pages        = {625--630},
	year         = 2024
}

@inproceedings{jiang-etal-2024-large,
	title        = {On Large Language Models' Hallucination with Regard to Known Facts},
	author       = {Jiang, Che  and Qi, Biqing  and Hong, Xiangyu  and Fu, Dayuan  and Cheng, Yang  and Meng, Fandong  and Yu, Mo  and Zhou, Bowen  and Zhou, Jie},
	booktitle    = {Proceedings of the 2024 Conference of the North American Chapter of the Association for Computational Linguistics: Human Language Technologies (Volume 1: Long Papers)},
	year         = 2024,
	pages        = {1041--1053}
}

@inproceedings{pmlr-v235-park24c,
	title        = {The Linear Representation Hypothesis and the Geometry of Large Language Models},
	author       = {Park, Kiho and Choe, Yo Joong and Veitch, Victor},
	booktitle    = {Proceedings of the 41st International Conference on Machine Learning},
	pages        = {39643--39666},
	year         = 2024,
	volume       = 235
}

@inproceedings{bao-etal-2025-probing,
	title        = {Probing the Geometry of Truth: Consistency and Generalization of Truth Directions in {LLM}s Across Logical Transformations and Question Answering Tasks},
	author       = {Bao, Yuntai  and Zhang, Xuhong  and Du, Tianyu  and Zhao, Xinkui  and Feng, Zhengwen  and Peng, Hao  and Yin, Jianwei},
	booktitle    = {Findings of the Association for Computational Linguistics: ACL 2025},
	year         = 2025,
	pages        = {682--700},
	isbn         = {979-8-89176-256-5}
}

@article{Lei03072018,
	author       = {Jing Lei and Max G’Sell and Alessandro Rinaldo and Ryan J. Tibshirani and Larry Wasserman},
	title        = {Distribution-Free Predictive Inference for Regression},
	journal      = {Journal of the American Statistical Association},
	volume       = 113,
	number       = 523,
	pages        = {1094--1111},
	year         = 2018
}

@article{gibbs2025conformal,
	title        = {Conformal prediction with conditional guarantees},
	author       = {Gibbs, Isaac and Cherian, John J and Cand{\`e}s, Emmanuel J},
	journal      = {Journal of the Royal Statistical Society Series B: Statistical Methodology},
	volume       = 87,
	number       = 4,
	pages        = {1100--1126},
	year         = 2025
}

@inproceedings{pmlr-v204-kato23a,
	title        = {A Review of Nonconformity Measures for Conformal Prediction in Regression},
	author       = {Kato, Yuko and Tax, David M.J. and Loog, Marco},
	booktitle    = {Proceedings of the Twelfth Symposium on Conformal and Probabilistic Prediction with Applications},
	pages        = {369--383},
	year         = 2023,
	volume       = 204
}

@article{shafer2011test,
	title        = {Test martingales, \text{Bayes} factors and p-values},
	author       = {Shafer, Glenn and Shen, Alexander and Vereshchagin, Nikolai and Vovk, Vladimir},
	journal      = {Statistical Science},
	pages        = {84--101},
	year         = 2011
}

@article{howard2020time,
	author       = {Howard, Steven and Ramdas, Aaditya and McAuliffe, Jon and Sekhon, Jagmohan},
	year         = 2020,
	pages        = {257--317},
	title        = {Time-uniform \text{Chernoff} bounds via nonnegative supermartingales},
	volume       = 17,
	journal      = {Probability Surveys}
}

@article{grunwald2024beyond,
	title        = {Beyond Neyman--Pearson: E-values enable hypothesis testing with a data-driven alpha},
	author       = {Gr{\"u}nwald, Peter D},
	journal      = {Proceedings of the National Academy of Sciences},
	volume       = 121,
	number       = 39,
	pages        = {e2302098121},
	year         = 2024
}

@inproceedings{hendrycks2017a,
	title        = {A Baseline for Detecting Misclassified and Out-of-Distribution Examples in Neural Networks},
	author       = {Dan Hendrycks and Kevin Gimpel},
	booktitle    = {International Conference on Learning Representations},
	year         = 2017
}

@inproceedings{NEURIPS2018_abdeb6f5,
	author       = {Lee, Kimin and Lee, Kibok and Lee, Honglak and Shin, Jinwoo},
	booktitle    = {Advances in Neural Information Processing Systems},
	title        = {A Simple Unified Framework for Detecting Out-of-Distribution Samples and Adversarial Attacks},
	volume       = 31,
	year         = 2018
}

@inproceedings{NEURIPS2020_f5496252,
	author       = {Liu, Weitang and Wang, Xiaoyun and Owens, John and Li, Yixuan},
	booktitle    = {Advances in Neural Information Processing Systems},
	pages        = {21464--21475},
	title        = {Energy-based Out-of-distribution Detection},
	volume       = 33,
	year         = 2020
}

@misc{touvron2023llama2openfoundation,
      title={Llama 2: Open Foundation and Fine-Tuned Chat Models}, 
      author={Hugo Touvron and Louis Martin and Kevin Stone and Peter Albert and Amjad Almahairi and Yasmine Babaei and Nikolay Bashlykov and Soumya Batra and Prajjwal Bhargava and Shruti Bhosale and Dan Bikel and Lukas Blecher and Cristian Canton Ferrer and Moya Chen and Guillem Cucurull and David Esiobu and Jude Fernandes and Jeremy Fu and Wenyin Fu and Brian Fuller and Cynthia Gao and Vedanuj Goswami and Naman Goyal and Anthony Hartshorn and Saghar Hosseini and Rui Hou and Hakan Inan and Marcin Kardas and Viktor Kerkez and Madian Khabsa and Isabel Kloumann and Artem Korenev and Punit Singh Koura and Marie-Anne Lachaux and Thibaut Lavril and Jenya Lee and Diana Liskovich and Yinghai Lu and Yuning Mao and Xavier Martinet and Todor Mihaylov and Pushkar Mishra and Igor Molybog and Yixin Nie and Andrew Poulton and Jeremy Reizenstein and Rashi Rungta and Kalyan Saladi and Alan Schelten and Ruan Silva and Eric Michael Smith and Ranjan Subramanian and Xiaoqing Ellen Tan and Binh Tang and Ross Taylor and Adina Williams and Jian Xiang Kuan and Puxin Xu and Zheng Yan and Iliyan Zarov and Yuchen Zhang and Angela Fan and Melanie Kambadur and Sharan Narang and Aurelien Rodriguez and Robert Stojnic and Sergey Edunov and Thomas Scialom},
      year={2023},
      eprint={2307.09288},
      archivePrefix={arXiv},
      primaryClass={cs.CL},
      url={https://arxiv.org/abs/2307.09288}, 
}

@misc{qwen2025qwen25technicalreport,
      title={Qwen2.5 Technical Report}, 
      author={An Yang and Baosong Yang and Beichen Zhang and Binyuan Hui and Bo Zheng and Bowen Yu and Chengyuan Li and Dayiheng Liu and Fei Huang and Haoran Wei and Huan Lin and Jian Yang and Jianhong Tu and Jianwei Zhang and Jianxin Yang and Jiaxi Yang and Jingren Zhou and Junyang Lin and Kai Dang and Keming Lu and Keqin Bao and Kexin Yang and Le Yu and Mei Li and Mingfeng Xue and Pei Zhang and Qin Zhu and Rui Men and Runji Lin and Tianhao Li and Tianyi Tang and Tingyu Xia and Xingzhang Ren and Xuancheng Ren and Yang Fan and Yang Su and Yichang Zhang and Yu Wan and Yuqiong Liu and Zeyu Cui and Zhenru Zhang and Zihan Qiu},
      year={2025},
      eprint={2412.15115},
      archivePrefix={arXiv},
      primaryClass={cs.CL},
      url={https://arxiv.org/abs/2412.15115}, 
}

@misc{novello2024outofdistributiondetectionuseconformal,
      title={Out-of-Distribution Detection Should Use Conformal Prediction (and Vice-versa?)}, 
      author={Paul Novello and Joseba Dalmau and Léo Andeol},
      year={2024},
      eprint={2403.11532},
      archivePrefix={arXiv},
      primaryClass={stat.ML},
      url={https://arxiv.org/abs/2403.11532}, 
}

@inproceedings{gupta-etal-2025-polysemantic,
	title        = {Polysemantic Dropout: Conformal \text{OOD} Detection for Specialized LLMs},
	author       = {Gupta, Ayush and Kaur, Ramneet and Roy, Anirban and Cobb, Adam D. and Chellappa, Rama and Jha, Susmit},
	booktitle    = {Proceedings of the 2025 Conference on Empirical Methods in Natural Language Processing},
	year         = 2025,
	pages        = {11757--11770}
}

@inproceedings{liu-etal-2024-good,
	title        = {How Good Are {LLM}s at Out-of-Distribution Detection?},
	author       = {Liu, Bo  and Zhan, Li-Ming  and Lu, Zexin  and Feng, Yujie  and Xue, Lei  and Wu, Xiao-Ming},
	booktitle    = {Proceedings of the 2024 Joint International Conference on Computational Linguistics, Language Resources and Evaluation},
	year         = 2024,
	pages        = {8211--8222}
}

@inproceedings{zhao2021detecting,
  title={Detecting operational adversarial examples for reliable deep learning},
  author={Zhao, Xingyu and Huang, Wei and Schewe, Sven and Dong, Yi and Huang, Xiaowei},
  booktitle={2021 51st Annual IEEE/IFIP International Conference on Dependable Systems and Networks-Supplemental Volume (DSN-S)},
  pages={5--6},
  year={2021},
  organization={IEEE}
}

@article{dong2023reliability,
  title={Reliability assessment and safety arguments for machine learning components in system assurance},
  author={Dong, Yi and Huang, Wei and Bharti, Vibhav and Cox, Victoria and Banks, Alec and Wang, Sen and Zhao, Xingyu and Schewe, Sven and Huang, Xiaowei},
  journal={ACM transactions on embedded computing systems},
  volume={22},
  number={3},
  pages={1--48},
  year={2023},
  publisher={ACM New York, NY}
}

@INPROCEEDINGS{jiang2025llmprism,
  author={Jiang, Zhihan and Ren, Rui and Yu, Guangba and Wu, Yulun and Gu, Wenwei and Li, Yichen and Huang, Yujie and Feng, Cong and Yang, Zengyin and Yang, Yongqiang and Lyu, Michael R.},
  booktitle={2025 55th Annual IEEE/IFIP International Conference on Dependable Systems and Networks - Supplemental Volume (DSN-S)}, 
  title={LLMPrism: Black-box Performance Diagnosis for Production LLM Training Platforms}, 
  year={2025},
  volume={},
  number={},
  pages={1-7},
  doi={10.1109/DSN-S65789.2025.00034}}

@article{li2025sr,
  title={FragileFlow: Spectral Control of Correct-but-Fragile Predictions for Foundation Model Robustness},
  author={Li, Zhuoyun and Wang, Boxuan and Hu, Jinwei and Huang, Xiaowei and Dong, Yi},
  journal={arXiv preprint arXiv:2605.08896},
  year={2026}
}

@misc{jiang2023mistral7b,
  title         = {Mistral 7B},
  author        = {Jiang, Albert Q. and Sablayrolles, Alexandre and Mensch, Arthur and Bamford, Chris and Chaplot, Devendra Singh and Casas, Diego de las and Bressand, Florian and Lengyel, Gianna and Lample, Guillaume and Saulnier, Lucile and others},
  year          = {2023},
  eprint        = {2310.06825},
  archivePrefix = {arXiv},
  primaryClass  = {cs.CL},
  doi           = {10.48550/arXiv.2310.06825},
  url           = {https://arxiv.org/abs/2310.06825}
}

@misc{olmo2025olmo2,
  title         = {2 OLMo 2 Furious},
  author        = {{Team OLMo} and Walsh, Pete and Soldaini, Luca and Groeneveld, Dirk and Lo, Kyle and Arora, Shane and Bhagia, Akshita and Gu, Yuling and Huang, Shengyi and Jordan, Matt and Lambert, Nathan and others},
  year          = {2025},
  eprint        = {2501.00656},
  archivePrefix = {arXiv},
  primaryClass  = {cs.CL},
  doi           = {10.48550/arXiv.2501.00656},
  url           = {https://arxiv.org/abs/2501.00656}
}

@misc{almazrouei2023falcon,
  title         = {The Falcon Series of Open Language Models},
  author        = {Almazrouei, Ebtesam and Alobeidli, Hamza and Alshamsi, Abdulaziz and Cappelli, Alessandro and Cojocaru, Ruxandra and Debbah, M{\'e}rouane and Goffinet, {\'E}tienne and Hesslow, Daniel and Launay, Julien and Malartic, Quentin and Mazzotta, Daniele and Noune, Badreddine and Pannier, Baptiste and Penedo, Guilherme},
  year          = {2023},
  eprint        = {2311.16867},
  archivePrefix = {arXiv},
  primaryClass  = {cs.CL},
  doi           = {10.48550/arXiv.2311.16867},
  url           = {https://arxiv.org/abs/2311.16867}
}

@inproceedings{socher2013recursive,
  title     = {Recursive Deep Models for Semantic Compositionality Over a Sentiment Treebank},
  author    = {Socher, Richard and Perelygin, Alex and Wu, Jean and Chuang, Jason and Manning, Christopher D. and Ng, Andrew and Potts, Christopher},
  booktitle = {Proceedings of the 2013 Conference on Empirical Methods in Natural Language Processing},
  pages     = {1631--1642},
  year      = {2013},
  address   = {Seattle, Washington, USA},
  publisher = {Association for Computational Linguistics},
  url       = {https://aclanthology.org/D13-1170/}
}

@inproceedings{wang2018glue,
  title     = {{GLUE}: A Multi-Task Benchmark and Analysis Platform for Natural Language Understanding},
  author    = {Wang, Alex and Singh, Amanpreet and Michael, Julian and Hill, Felix and Levy, Omer and Bowman, Samuel R.},
  booktitle = {Proceedings of the 2018 EMNLP Workshop BlackboxNLP: Analyzing and Interpreting Neural Networks for NLP},
  pages     = {353--355},
  year      = {2018},
  address   = {Brussels, Belgium},
  publisher = {Association for Computational Linguistics},
  doi       = {10.18653/v1/W18-5446},
  url       = {https://aclanthology.org/W18-5446/}
}

@inproceedings{lang1995newsweeder,
  title     = {NewsWeeder: Learning to Filter Netnews},
  author    = {Lang, Ken},
  booktitle = {Machine Learning Proceedings 1995},
  pages     = {331--339},
  year      = {1995},
  publisher = {Elsevier},
  doi       = {10.1016/B978-1-55860-377-6.50048-7}
}

@inproceedings{larson2019evaluation,
  title     = {An Evaluation Dataset for Intent Classification and Out-of-Scope Prediction},
  author    = {Larson, Stefan and Mahendran, Anish and Peper, Joseph J. and Clarke, Christopher and Lee, Andrew and Hill, Parker and Kummerfeld, Jonathan K. and Leach, Kevin and Laurenzano, Michael A. and Tang, Lingjia and Mars, Jason},
  booktitle = {Proceedings of the 2019 Conference on Empirical Methods in Natural Language Processing and the 9th International Joint Conference on Natural Language Processing},
  pages     = {1311--1316},
  year      = {2019},
  address   = {Hong Kong, China},
  publisher = {Association for Computational Linguistics},
  doi       = {10.18653/v1/D19-1131},
  url       = {https://aclanthology.org/D19-1131/}
}

@inproceedings{casanueva2020efficient,
  title     = {Efficient Intent Detection with Dual Sentence Encoders},
  author    = {Casanueva, I{\~n}igo and Tem{\v{c}}inas, Tadas and Gerz, Daniela and Henderson, Matthew and Vuli{\'c}, Ivan},
  booktitle = {Proceedings of the 2nd Workshop on Natural Language Processing for Conversational AI},
  pages     = {38--45},
  year      = {2020},
  address   = {Online},
  publisher = {Association for Computational Linguistics},
  doi       = {10.18653/v1/2020.nlp4convai-1.5},
  url       = {https://aclanthology.org/2020.nlp4convai-1.5/}
}

@incollection{dagan2006pascal,
  title     = {The {PASCAL} Recognising Textual Entailment Challenge},
  author    = {Dagan, Ido and Glickman, Oren and Magnini, Bernardo},
  booktitle = {Machine Learning Challenges: Evaluating Predictive Uncertainty, Visual Object Classification, and Recognising Textual Entailment},
  pages     = {177--190},
  year      = {2006},
  publisher = {Springer},
  series    = {Lecture Notes in Computer Science},
  volume    = {3944},
  doi       = {10.1007/11736790_9}
}

@inproceedings{bojar2014findings,
  title     = {Findings of the 2014 Workshop on Statistical Machine Translation},
  author    = {Bojar, Ond{\v{r}}ej and Buck, Christian and Federmann, Christian and Haddow, Barry and Koehn, Philipp and Leveling, Johannes and Monz, Christof and Pecina, Pavel and Post, Matt and Saint-Amand, Herv{\'e} and Soricut, Radu and Specia, Lucia and Tamchyna, Ale{\v{s}}},
  booktitle = {Proceedings of the Ninth Workshop on Statistical Machine Translation},
  pages     = {12--58},
  year      = {2014},
  address   = {Baltimore, Maryland, USA},
  publisher = {Association for Computational Linguistics},
  url       = {https://aclanthology.org/W14-3302/}
}

@inproceedings{li2002learning,
  title     = {Learning Question Classifiers},
  author    = {Li, Xin and Roth, Dan},
  booktitle = {{COLING} 2002: The 19th International Conference on Computational Linguistics},
  year      = {2002},
  url       = {https://aclanthology.org/C02-1150/}
}

@inproceedings{williams2018broad,
  title     = {A Broad-Coverage Challenge Corpus for Sentence Understanding through Inference},
  author    = {Williams, Adina and Nangia, Nikita and Bowman, Samuel},
  booktitle = {Proceedings of the 2018 Conference of the North American Chapter of the Association for Computational Linguistics: Human Language Technologies, Volume 1},
  pages     = {1112--1122},
  year      = {2018},
  address   = {New Orleans, Louisiana},
  publisher = {Association for Computational Linguistics},
  doi       = {10.18653/v1/N18-1101},
  url       = {https://aclanthology.org/N18-1101/}
}
